\documentclass[]{fairmeta}

\usepackage{xcolor}
\usepackage{tgpagella}
\usepackage{bbding}

\usepackage{graphicx}
\usepackage{booktabs}
\usepackage{amsmath}
\usepackage{amssymb}
\usepackage{multirow}
\usepackage{pifont}
\usepackage{soul}
\usepackage{lipsum} 
\usepackage{bm}
\usepackage[position=top]{subcaption}
\usepackage{colortbl}
\usepackage{algorithm}
\usepackage{algpseudocode}
\usepackage{tabu}
\usepackage{arydshln}
\usepackage{wrapfig}

\renewcommand{\paragraph}[1]{\noindent\textbf{#1}}

\definecolor{pos}{RGB}{0,153,0}
\definecolor{neg}{RGB}{0,0,0}
\definecolor{smpos}{RGB}{0,0,0}
\definecolor{rowx}{RGB}{242, 243, 244}
\definecolor{row}{RGB}{235, 245, 251}
\definecolor{suppcolor}{RGB}{0,0,255}
\definecolor{down}{RGB}{153, 163, 164}
\definecolor{down2}{RGB}{153, 204, 205}

\RequirePackage{xspace}
\makeatletter
\DeclareRobustCommand\onedot{\futurelet\@let@token\@onedot}
\def\@onedot{\ifx\@let@token.\else.\null\fi\xspace}
\def\eg{\emph{e.g}\onedot} 

\def\ie{\emph{i.e}\onedot}

\makeatother

\newcommand{\cmark}{\ding{51}}%
\newcommand{\xmark}{\ding{55}}%
\newcommand{\fref}[1]{Fig.~\ref{#1}}
\newcommand{\tref}[1]{Table~\ref{#1}}

\DeclareCaptionLabelFormat{andtable}{#1~#2  \&  \tablename~\thetable}

\title{Adaptive Caching for Faster Video Generation \\with Diffusion Transformers}
\author[1,2,*]{Kumara Kahatapitiya}
\author[1]{Haozhe Liu}
\author[1]{Sen He}
\author[1]{Ding Liu}
\author[1]{Menglin Jia}
\author[1]{Chenyang Zhang}
\author[2]{Michael S. Ryoo}
\author[1]{Tian Xie}
\affiliation[1]{Meta AI}
\affiliation[2]{Stony Brook University}

\contribution[*]{Work done at Meta}

\abstract{
\begin{abstract}

Generating temporally-consistent high-fidelity videos can be computationally expensive, especially over longer temporal spans. More-recent Diffusion Transformers (DiTs)--- despite making significant headway in this context--- have only heightened such challenges as they rely on larger models and heavier attention mechanisms, resulting in slower inference speeds. In this paper, we introduce a \textit{training-free} method to accelerate video DiTs, termed Adaptive Caching (\textit{AdaCache}), which is motivated by the fact that \textit{``not all videos are created equal''}: meaning, some videos require fewer denoising steps to attain a reasonable quality than others. Building on this, we not only cache computations through the diffusion process, but also devise a caching schedule tailored to each video generation, maximizing the quality-latency trade-off. 
We further introduce a Motion Regularization (\textit{MoReg}) scheme 
to utilize video information within AdaCache, essentially controlling the compute allocation based on motion content. Altogether, our plug-and-play contributions grant significant inference speedups (\eg up to 4.7$\times$ on Open-Sora 720p - 2s video generation) without sacrificing the generation quality, across multiple video DiT baselines.

\end{abstract}
}

\date{\today}
\correspondence{Kumara Kahatapitiya at \email{kkahatapitiy@cs.stonybrook.edu}, Tian Xie at \email{tianxie@meta.com}}

\metadata[Blogpost]{\url{https://adacache-dit.github.io}}

\begin{document}

\maketitle

\section{Introduction}
\label{sec:intro}

Diffusion models \citep{ho2020ddpm, song2020ddim} have become the standard for generative modeling in recent years, arguably surpassing the quality of VAEs \citep{kingma2013vae, rolfe2016discretevae}, GANs \citep{karras2019stylegan, goodfellow2020gan} and Auto-Regressive models \citep{chang2022maskgit, chang2023muse}. This observation holds in a wide-range of applications including image \citep{rombach2022stablediffusion, saharia2022imagen}, video \citep{singer2022makeavideo, blattmann2023svd}, 3D \citep{poole2022dreamfusion, liu2023zero123}, and audio \citep{kong2020diffwave, huang2023makeanaudio} generation, as well as image \citep{hertz2022p2p, avrahami2023blended} and video \citep{qi2023fatezero, wu2023tuneavideo} editing. More recent Diffusion Transformers (DiTs) \citep{peebles2023dit, ma2024sit} show better promise in terms of scalability and generalization 
compared to prior UNet-based diffusion models \citep{rombach2022stablediffusion}, revealing intriguing horizons in GenAI for the years to come. 

Despite the state-of-the-art performance, DiTs can also be computationally expensive both in terms of memory and computational requirements. This becomes especially critical when applied with a large number of input tokens (\eg high-resolution long video generation). For instance, the reason for models such as Sora \citep{brooks2024sora} not being publicly-served is speculated to be the high resource demands and slower inference speeds \citep{liu2024sorareview}. To tackle these challenges and reduce the footprint of diffusion models, various research directions have emerged such as latent diffusion \citep{rombach2022stablediffusion}, step-distillation \citep{sauer2023adversarial, yin2024one}, caching \citep{wimbauer2024blockcache, ma2024deepcache, habibian2024clockwork}, architecture-search \citep{zhao2023mobilediffusion, li2024snapfusion}, token reduction \citep{bolya2023tomesd, li2024vidtome} and region-based methods \citep{nitzan2024lazydiff, kahatapitiya2024ocd}. Fewer techniques transfer readily from UNet-based pipelines to DiTs, whereas others often require novel formulations. Hence, DiT acceleration has been under-explored as of yet.

\begin{figure}[t]
\centering
\includegraphics[width=1\columnwidth]{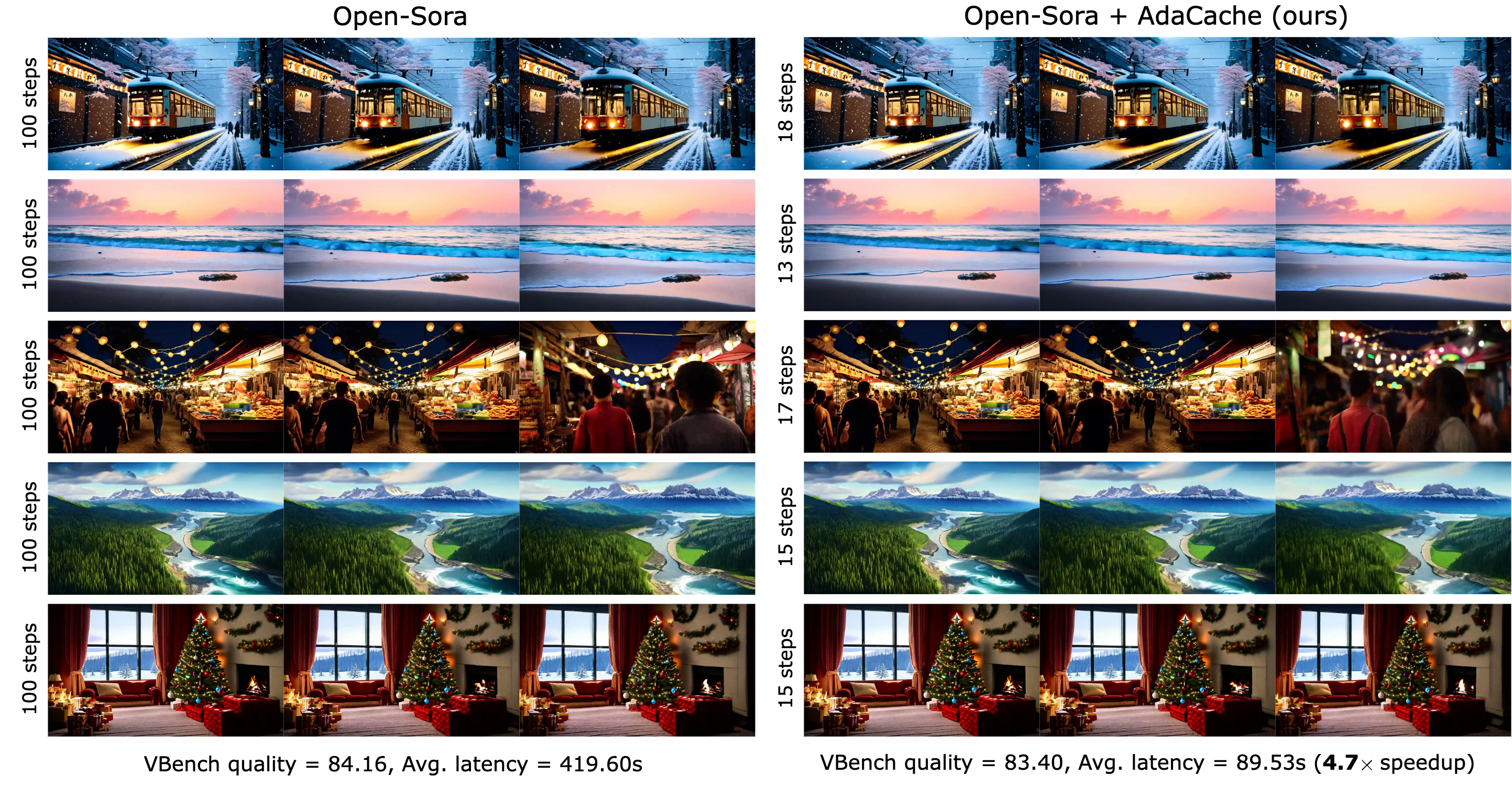}
\vspace{-8mm}
\caption{\textbf{Effectiveness of Adaptive Caching:} We show a qualitative comparison of AdaCache (right) applied on top of Open-Sora \citep{opensora} (left), a baseline video DiT. Here, we consider generating 720p - 2s video clips, and report VBench \citep{huang2023vbench} quality and average latency (on a single A100 GPU) on the benchmark prompts from Open-Sora gallery. AdaCache generates videos significantly faster (\ie, 4.7$\times$ speedup) with a comparable quality. Also, the number of computed steps varies for each video. Best-viewed with zoom-in. Prompts given in supplementary.
}
\label{fig:teaser}
\end{figure}

Moreover, we note that \textit{not all videos are created equal}. Some videos contain high-frequency textures and significant motion content, whereas others are much simpler (\eg with homogeneous textures or static regions). Having a diffusion process tailored specifically for each video generation can be beneficial in terms of realizing the best quality-latency trade-off. This idea has been explored to some extent in region-based methods \citep{avrahami2023blended, nitzan2024lazydiff, kahatapitiya2024ocd}, but not sufficiently in the context of video generation.

Motivated by the above, we introduce Adaptive Caching (\textit{AdaCache}) for accelerating video diffusion transformers. This approach requires \textit{no training} and can seamlessly be integrated into a baseline video DiT at inference, as a plug-and-play component. The core idea of our proposal is to cache residual computations within transformer blocks (\eg attention or MLP outputs) in a certain diffusion step, and reuse them through a number of subsequent steps that is dependent on the generated video. We do this by devising a caching schedule, \ie, deciding when-to-recompute-next whenever making a residual computation. This decision is guided by a distance metric that measures the rate-of-change between previously-stored and current representations. If the distance is high we would not cache for an extended period (\ie, \#steps), to avoid reusing incompatible representations. We further introduce a Motion Regularization (\textit{MoReg}) to allocate computations based on the motion content in the video being generated. This is inspired by the observation that high-moving sequences require more diffusion steps to achieve a reasonable quality.
Altogether, our pipeline is applied on top of multiple video DiT baselines showing much-faster inference speeds without sacrificing the generation quality (see \fref{fig:teaser}). Finally, we validate the effectiveness of our contributions and justify our design decisions through ablations and qualitative comparisons.
\section{Related Work}

\paragraph{Diffusion-based Video Generation} \citep{singer2022makeavideo, ho2022imagenvideo, blattmann2023svd, girdhar2023emuvideo, chen2024videocrafter2} has surpassed the quality and diversity of GAN-based approaches \citep{vondrick2016generating, saito2017temporal, tulyakov2018mocogan, clark2019adversarial, yu2022generating}, while also being competitive with recent Auto-Regressive models \citep{yan2021videogpt, hong2022cogvideo, villegas2022phenaki, kondratyuk2023videopoet, xie2024show, liu2024mardini}. They have become a standard component in the pipelines for frame interpolation \citep{wang2024generativeinbetweening, feng2024explorative}, video outpainting \citep{fan2023hierarchical, chen2024followyourcanvas, wang2024beyouroutpainter}, image-to-video \citep{guo2023i2v, blattmann2023svd, xing2023dynamicrafter},  video-to-video (\ie, video editing or translation) \citep{yang2023rerender, yatim2024smm, hu2024depthcrafter}, personalization \citep{wu2024customcrafter, men2024mimo}, motion customization \citep{zhao2023motiondirector, xu2024camco} and compositional generation \citep{liu2022compositional, yang2024compositional}. The underlying architecture of video diffusion models has evolved from classical UNets \citep{ronneberger2015unet, rombach2022stablediffusion} with additional spatio-temporal attention layers \citep{he2022latentvideodm, blattmann2023align, chen2023seine, girdhar2023emuvideo}, to fully-fledged transformer-based (\ie, DiT \citep{peebles2023dit}) architectures \citep{lu2023vdt, ma2024latte, gao2024lumina, zhang2024tora}. In the process, the latency of denoising \citep{song2020ddim, lu2022dpm} has also scaled with larger models \citep{podell2023sdxl, gao2024lumina}. This becomes critical especially in applications such as long-video generation \citep{yin2023nuwa, wang2023genlvideo, zhao2024moviedreamer, henschel2024streamingt2v, tan2024videoinfinity, zhou2024storydiffusion}, while also affecting the growth of commercially-served video models \citep{gen3, brooks2024sora, luma, kling}.

\paragraph{Efficiency of Diffusion models} has been actively explored with respect to both training and inference pipelines. Multi-stage training at varying resolutions \citep{chen2023pixartalpha, chen2024pixartsigma, gao2024lumina} and high-quality data curation \citep{ramesh2022dalle2, ho2022imagenvideo, dai2023emu, blattmann2023svd} have cut down training costs significantly. In terms of inference acceleration, there exist two main approaches: (1) methods that require re-training such as step-distillation \citep{salimans2022progressivedistillation, meng2023guideddistillation, sauer2023adversarial, liu2023instaflow}, consistency regularization \citep{song2023consistency, luo2023latentconsistency}, quantization \citep{li2023qdiff, chen2024qdit, he2024ptqd, wang2024quest, deng2024vq4dit}, and architecture search/compression \citep{zhao2023mobilediffusion, yang2023slimdiffusion, li2024snapfusion}, or (2) methods that require no re-training such as token reduction \citep{bolya2023tomesd, li2024vidtome, kahatapitiya2024ocd} and caching \citep{ma2024deepcache, wimbauer2024blockcache, habibian2024clockwork, chen2024deltadit, zhao2024pab}. Among these, training-free methods are more-attractive as they can be widely-adopted without any additional costs. This becomes especially relevant for video diffusion models that are both expensive to train and usually very slow at inference. In this paper, we explore a caching-based approach tailored for video DiTs. Different from prior fixed caching schedules in UNet-based \citep{ma2024deepcache, wimbauer2024blockcache, habibian2024clockwork} and DiT-based \citep{chen2024deltadit, zhao2024pab} pipelines, we introduce a content-dependent (\ie, adaptive) caching scheme to squeeze out the best quality-latency trade-off.

\paragraph{Content-adaptive Generation} may focus on improving consistency \citep{couairon2022diffedit, bar2022text2live, avrahami2022blended, avrahami2023blended, wang2023imageneditor, xie2023smartbrush}, quality \citep{suin2024diffuse, abu2022adir}, and/or efficiency \citep{tang2023deediff, nitzan2024lazydiff, kahatapitiya2024ocd, starodubcev2024adaptivediff}. Most region-based methods (\eg image or video editing) rely on a user-provided mask to ensue consistent generations aligned with context information \citep{avrahami2023blended, xie2023smartbrush}. Some others automatically detect \citep{suin2024diffuse} or retrieve \citep{abu2022adir} useful information to improve generation quality. Among efficiency-oriented approaches, there exist proposals for selectively-processing a subset of latents \citep{nitzan2024lazydiff, kahatapitiya2024ocd}, switching between diffusion models with varying compute budgets \citep{starodubcev2024adaptivediff}, or adaptively-controlling the number of denoising steps \citep{tang2023deediff, wimbauer2024blockcache}. AdaDiff \citep{tang2023deediff} skips all subsequent computations in a denoising step, if an uncertainty threshold is met at a certain layer. Block caching \citep{wimbauer2024blockcache} introduces a caching schedule tailored for a given pretrained diffusion model. Both these handle image generation tasks. In contrast, our proposed AdaCache--- which also controls \#denoising-steps adaptively--- provides better flexibility, and is applied to more-challenging video generation. It is flexible in the sense that (1) it can selectively-cache any layer or even just a specific module within a layer, and (2) it is tailored to each video generation instead of being fixed for a given architecture. Thus, AdaCache gains more control over the diffusion process, enabling a better-adaptive compute allocation.
\section{Not All Videos Are Created Equal}

In this section, we motivate the need for a content-dependent denoising process, and show how it can help maximize the quality-latency trade-off. This motivation is based on a couple of interesting observations which we describe below.

\begin{figure}[t!]
    \hspace{3mm}
     \begin{subfigure}{0.575\textwidth}
         \centering
         \includegraphics[width=1\columnwidth]{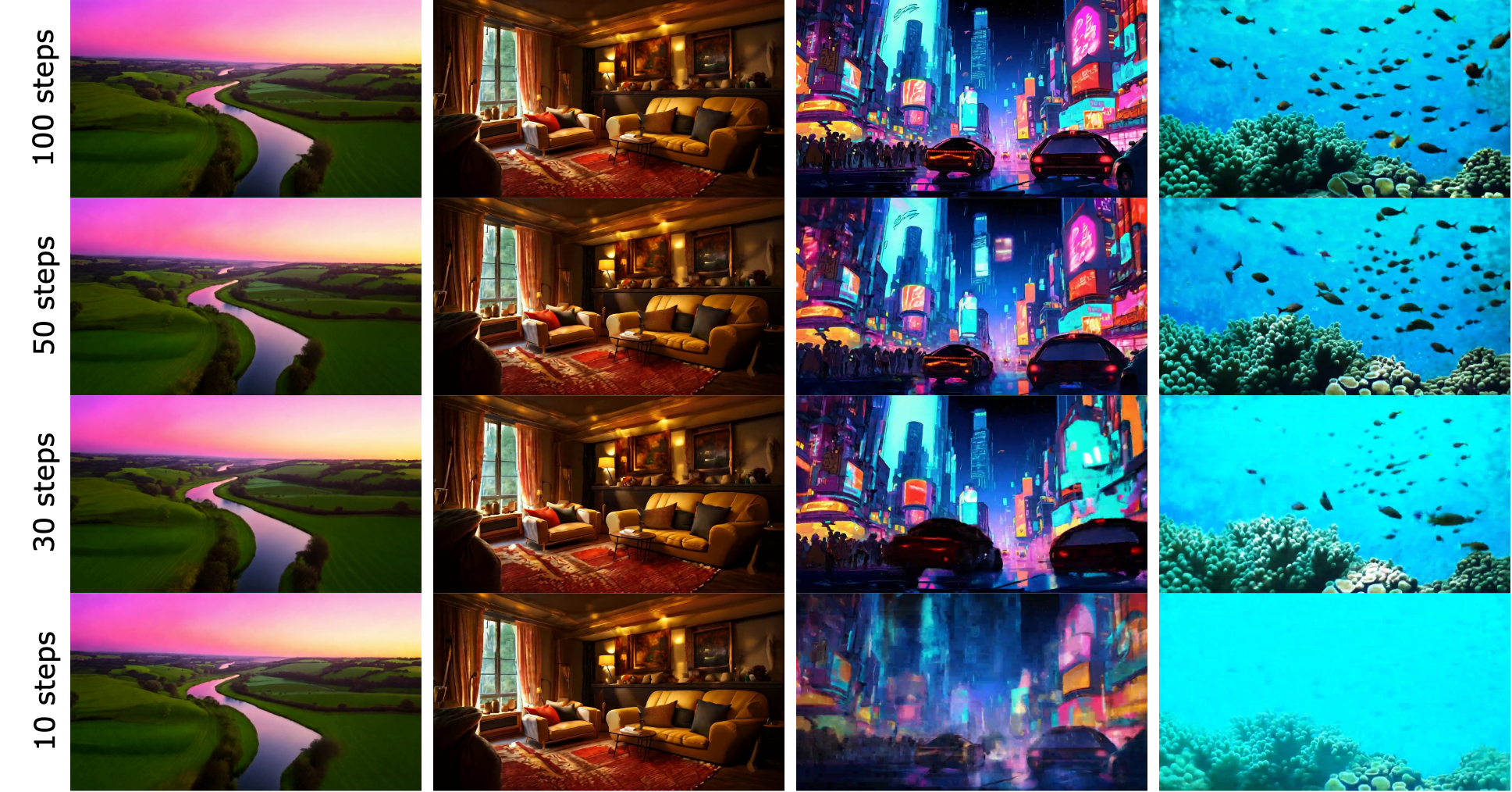}
     \end{subfigure}\hspace{1mm}
     \begin{subfigure}{0.385\textwidth}
         \centering
         \includegraphics[width=1\columnwidth]{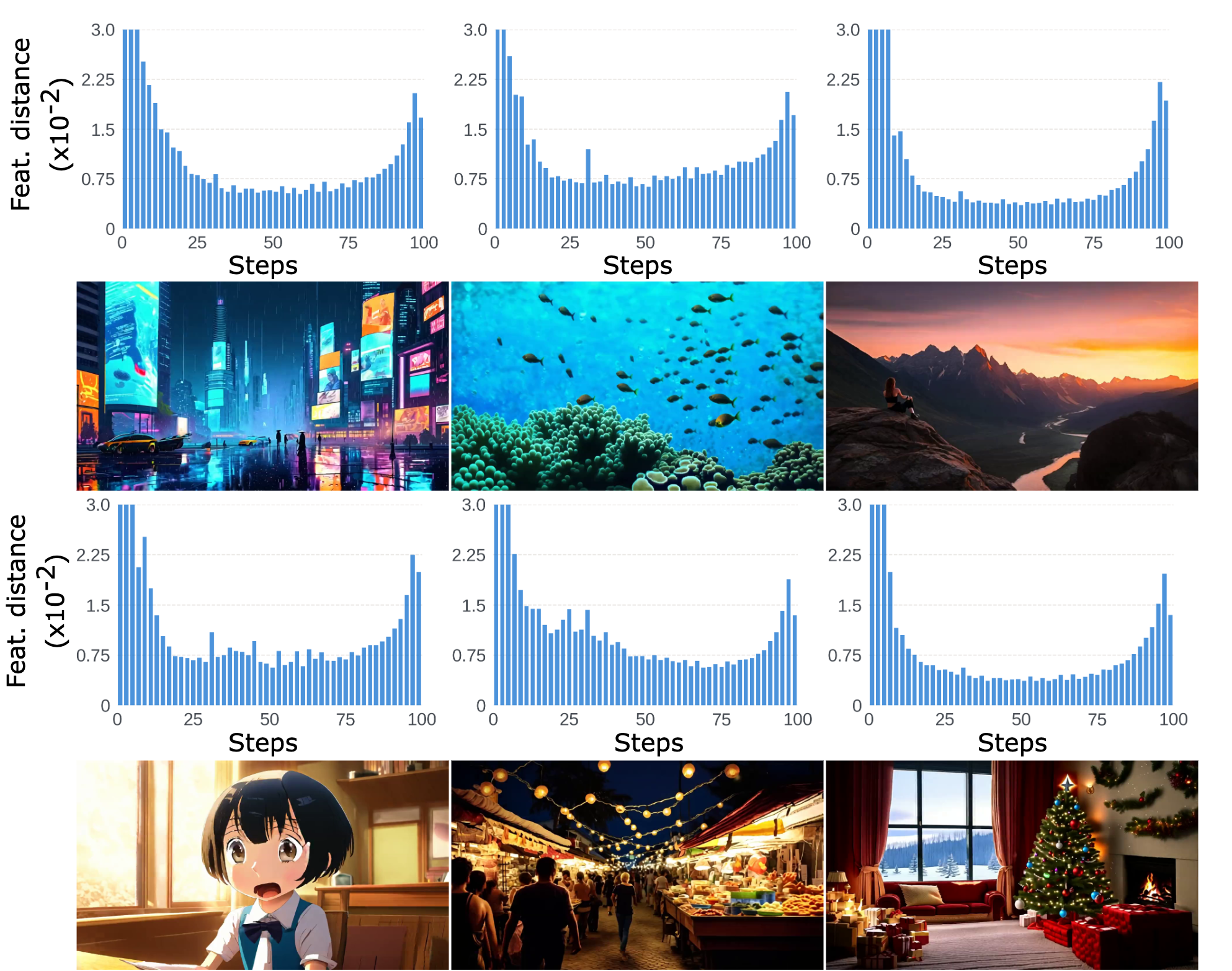}
     \end{subfigure}
     \caption{\textbf{Not all videos are created equal:} We show frames from 720p - 2s video generations based on Open-Sora \citep{opensora}. (Left) We try to break each generation by reducing the number of diffusion steps. Interestingly, not all videos have the same break point. Some sequences are extremely robust (\eg first-two columns), while others break easily. (Right) When we plot the difference between computed representations in subsequent diffusion steps, we see unique variations (Feature distance vs.~\#steps). If we are to reuse similar representations, it needs to be tailored to each video. Both these observations suggest the need for a content-dependent denoising process, which is the founding motivation of Adaptive Caching. Best-viewed with zoom-in. Prompts given in supplementary.}
     \vspace{-5mm}
     \label{fig:step_change}
\end{figure}

First, we note that each video is unique. Hence, videos have varying levels of complexity. Here, the complexity of a given video can be expressed by the rate-of-change of information across both space and time. Simpler videos may contain more homogeneous regions and/or static content. In contrast, complex videos have more high-frequency details and/or significant motion. The standard video compression techniques exploit such information to achieve the best possible compression ratios without sacrificing the quality \citep{wiegand2003h264, sullivan2012h265}. Motivated by the same, we explore how the compute-cost affects the quality of video generations based on DiTs. We measure this w.r.t.~the number of denoising steps, and the observations are shown in \fref{fig:step_change} (Left). Some video sequences are very robust, and achieve a reasonable quality even at fewer denoising steps. Others break easily when we keep reducing the \#steps, but the break-point varies. This observation suggests that the minimal \#steps (or, computations) required to generate a video with a reasonable quality varies, and having a content-dependent denoising schedule can exploit this to achieve the best speedups.

\begin{wrapfigure}{r}{0.475\textwidth}
\centering
\vspace{-4mm}
\includegraphics[width=0.475\columnwidth]{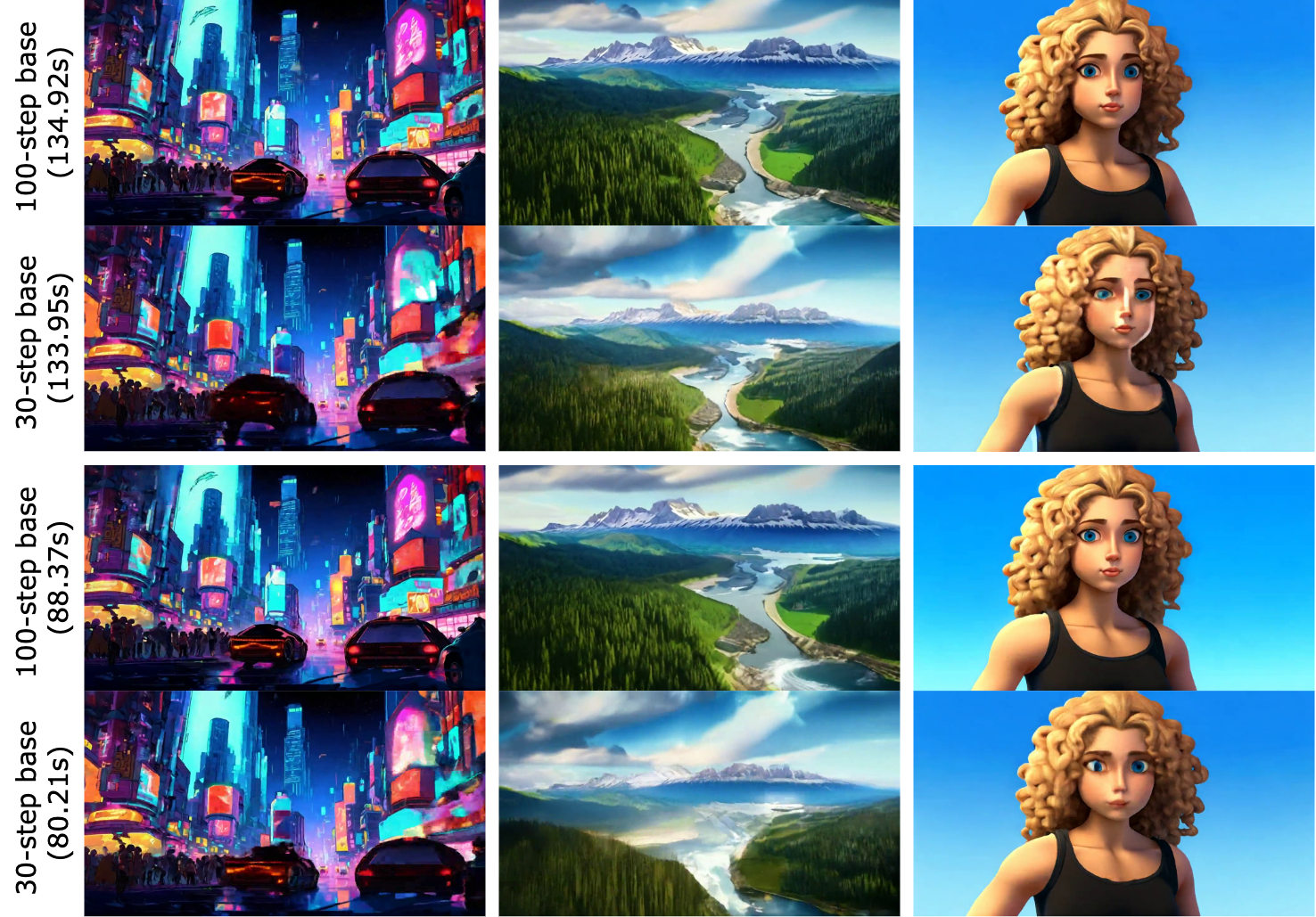}
\vspace{-5mm}
\caption{\textbf{Videos generated at a capped-budget:} There exist different configurations for generating videos at an approximately-fixed latency (\eg having an arbitrary \#denoising-steps, yet only computing a fixed \#representations and reusing otherwise). We observe a significant variance in quality in such videos. Best-viewed with zoom-in. Prompts given in supplementary.}
\vspace{-4mm}
\label{fig:30v100}
\end{wrapfigure}

Next, we observe how the computed representations (\ie, residual connections in attention or MLP blocks within DiT) change during the denoising process, across different video generations. This may reveal the level of compute redundancy in each video generation, enabling us to reuse representations and improve efficiency. More specifically, we visualize the feature differences between subsequent diffusion steps as histograms given in \fref{fig:step_change} (Right). Here, we report Feature distance (\eg L1) vs.~\#steps. We observe that each histogram is unique. Despite having higher changes in early/latter steps and smaller changes in the middle, the overall distribution and the absolute values vary considerably. A smaller change corresponds to higher redundancy across subsequent computations, and an opportunity for re-using. This motivates the need for a non-uniform compute-schedule not only within the diffusion process of a given video (\ie, at different stages of denoising), but also across different videos.

Finally, we evaluate the video generation quality at a capped-budget (\ie, fixed computations or latency). We can have multiple generation configurations at an approximately-fixed latency, by computing a constant number of representations. For instance, we can cache and reuse representations more-frequently in a setup with more denoising steps, still having the same latency of a process with fewer steps. The observations of a study with either 30 or 100 base denoising steps is shown in \fref{fig:30v100}. We see that the generation quality varies significantly despite spending a similar cost and having the same underlying pretrained DiT. This motivates us to think about how best to allocate our resources at inference, tailored for each video generation.

\section{Adaptive Caching for Faster Video DiTs}

\begin{figure}[h]
\centering
\includegraphics[width=0.8\columnwidth]{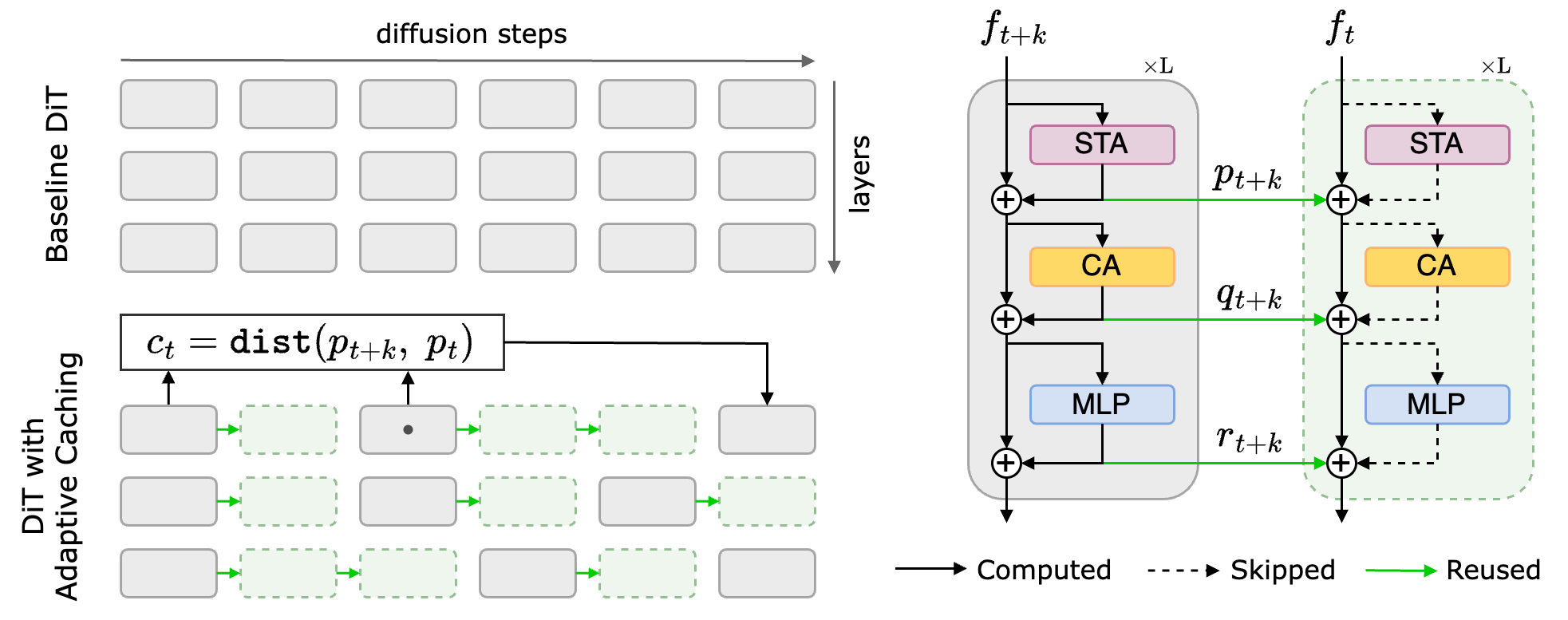}
\caption{\textbf{Overview of Adaptive Caching:} (Left) During the diffusion process, we choose to cache residual computations within selected DiT blocks. The caching schedule is \textit{content-dependent}, as we decide when to compute the next representation based on a distance metric ($c_t$). This metric measures the rate-of-change from previously-computed (and, stored) representation to the current one, and can be evaluated per-layer or the DiT as-a-whole. Each computed residual can be cached and reused across multiple steps. (Right) We only cache the residuals (\ie, skip-connections) which amount to the actual computations (\eg spatial-temporal/cross attention, MLP). The iteratively denoised representation (\ie, $f_{t+k}$, $f_{t}$) always gets updated either with computed or cached residuals.
}
\label{fig:main_fig}
\end{figure}

\subsection{Preliminaries: Video Diffusion Transformers}

Video Diffusion Transformers are extended from Latent Diffusion Transformers (DiTs) \citep{peebles2023dit} introduced for image generation. DiTs provide a much-more streamlined, scalable architecture compared to prior UNet-based diffusion models \citep{rombach2022stablediffusion}, by only having transformer blocks with a homogeneous token resolution (instead of convolutional blocks with up/downsampling). A simplified transformer block (\ie, w/o normalizing or timestep conditioning layers) in a video DiT is shown in \fref{fig:main_fig} (right)--- gray block. It consists of spatial-temporal attention (\texttt{STA}), cross-attention (\texttt{CA}) and linear (\texttt{MLP}) layers. Depending on the implementation, \texttt{STA} may be a single joint spatio-temporal attention layer, or separate spatial and temporal attention layers repeated within alternating blocks. Without loss of generality, let us denote a latent feature at the input/output of such block by $f^l_t$ and $f^{l+1}_t$, respectively. Here, $l$ represents the layer index, and $t$, the diffusion timestep. A simplified flow of computations within each block can be represented as,

\vspace{-5mm}
\begin{alignat}{2}
    p^l_t &= \texttt{STA}(f^l_t)\;&&;\quad \Tilde{f}^l_t = f^l_t + p^l_t\;, \\
    q^l_t &= \texttt{CA}(\Tilde{f}^l_t)\;&&;\quad \Bar{f}^l_t = \Tilde{f}^l_t + q^l_t\;, \\
    r^l_t &= \texttt{MLP}(\Bar{f}^l_t)\;&&;\quad f^{l+1}_t = \Bar{f}^l_t + r^l_t\;.
\end{alignat}

Here $p^l_t$, $q^l_t$ and $r^l_t$ are residual connections corresponding to each compute-element. Such computations repeat through $L$ layers, generating the noise prediction of each step $t$, and across a total of $T$ denoising steps. In the current streamlined video DiT architectures with homogeneous token resolutions, each layer of each denoising step costs the same.

\subsection{Adaptive Caching}

In this subsection, we introduce Adaptive Caching (\textit{AdaCache}), a \textit{training-free} mechanism for content-dependent compute allocation in video DiTs. The overview of Adaptive Caching is shown in \fref{fig:main_fig}. Compared to a standard DiT that computes representations for all layers across all diffusion steps, in AdaCache, we decide which layers or steps to compute, adaptively (\ie, dependent on each video being generated). This decision is based on the rate-of-change in the residual connections (\eg $p^l_t$, $q^l_t$ or $r^l_t$) across diffusion steps, which amount to all significant computations within the DiT. Without loss of generality, let us assume that the residuals in block $l$ in current and immediately-prior diffusion steps $t$ and $t+k$ are already computed. Here, step $t+k$ is identified as `immediately-prior' to step $t$ since any residuals between these two steps are assumed to be not computed (\ie, cached residuals are reused). We make a decision on the next computation step based on the distance metric ($c^l_t$) given by,

\vspace{-5mm}
\begin{alignat}{2}
    c^l_t &= \texttt{dist}(p^l_{t+k},\; p^l_{t}) = \|p^l_{t} - p^l_{t+k}\| \;/\;k\;.
\end{alignat}

Here, we use L1 distance by default, but other distance metrics can also be applied (\eg L2, cosine). Once we have the distance metric, we select the next caching rate ($\tau^l_t$) based on a pre-defined codebook of \textit{basis cache-rates}. Here, a `cache-rate' is defined as the number subsequent steps during which, a previously-computed representation is re-used (\ie, a higher cache-rate gives more compute savings). The codebook is basically a collection of cache-rates defined based on the original denoising schedule (\ie, \#steps), coupled with corresponding metric thresholds to select them. Simply put, a higher distance metric will sample a lower cache-rate from the codebook, resulting in more-frequent re-computations.

\vspace{-3mm}
\begin{alignat}{2}
    \tau^l_t &= \texttt{codebook}(c^l_t)\;.
\end{alignat}

For all denoising steps within $t$ and $t-\tau$, we reuse previously-cached representations and only recompute after the current caching schedule (while also estimating the metric, again).

\vspace{-3mm}
\begin{align}
p^l_{t-k} = 
\begin{cases}
\hspace{2mm}p^l_{t} \quad &\text{if} \quad k<\tau^l_t;\\
\hspace{2mm}p^l_{t-k} = \texttt{STA}(f^l_{t-k}) \quad &\text{if} \quad k=\tau^l_t.
\end{cases}
\end{align}

The same applies to other residual computations (\eg $q^l_{t-k}$, $r^l_{t-k}$) as well. By design, we can have unique caching schedules for each layer (and, each residual computation). However, we observe that it will make the generations unstable. Therefore, we decide to have a common metric (\ie, $c^l_t=c_t$) and hence, a common caching rate (\ie, $\tau^l_t=\tau_t$) across all DiT layers. For instance, we can consider an averaged metric across all layers, or a metric computed at a certain layer to decide the caching schedule. Meaning, when we recompute residuals in a certain step, we do so for the whole DiT rather than selectively for each layer.

Overall, this setup allows us to adaptively-control the compute spent on each video generation, based on frame-wise information (\ie, no temporal information used as of yet). If the rate-of-change between residuals is high, we will have a smaller caching rate, and otherwise, we have a higher rate. The choice of a lightweight distance metric (\eg L1) helps us avoid any additional latency overheads.

\subsection{Motion Regularization}

To further improve Adaptive Caching by making use of video-specific (\ie, temporal) information, we introduce a Motion Regularization (\textit{MoReg}). %
This is motivated by the observation that the optimal number of denoising steps varies based on the motion content of each generated video. The core idea here, is to cache less (\ie, recompute more) if a generated video has a high motion content. Simply put, we plan to regularize our caching schedule based on motion. However the problem is that, we need to estimate motion while the video is still being generated. Therefore, we can not rely on motion estimation algorithms in the pixel space, nor any compute-heavy ones as our focus is on efficiency. As a result, we estimate a \textit{noisy} latent motion-score ($m^l_t$) based on residual frame differences. Without loss of generality, let us denote residual latent frames of $p^l_t$ as $\{ p^l_{t,\;n}\;|\;n=0,\cdots,N-1\}$ where $N$ is the \#frames in latent space (generated by the VAE encoder). We estimate the motion-score as,

\vspace{-5mm}
\begin{alignat}{2}
    m^l_t &= \|p^l_{t,\;i:N} - p^l_{t,\;0:N-i}\| \;.
\end{alignat}

Here, $i$ denotes the frame step-size (or, frame-rate), $\|\cdot\|$, the distance metric (\eg L1), and $i:j$, the slice of all frames within the corresponding range. However, since we operate on noisy-latents, we observe that our motion estimate, particularly in early diffusion steps is not reliable. Meaning, it does not provide a reasonable regularization in early steps (\ie, the change in caching schedule does not correlate well with the observed motion in pixel space). To alleviate this, we also compute a motion-gradient ($mg^l_t$) across diffusion steps, which can act as a reasonable early-predictor of motion that we may observe in latter diffusion steps (that also correlates with the motion in pixel space).

\vspace{-3mm}
\begin{alignat}{2}
    mg^l_t &= (m^l_{t} - m^l_{t+k})\;/\; k\;.
\end{alignat}

Despite the motion-score being noisy, the motion-gradient acts as a better-estimate of trend, as the representations are getting denoised and converging to a noise-free distribution. Finally, we use both motion-score and motion-gradient as a scaling-factor of the distance metric ($c^l_t$) to regularize our caching schedule.

\vspace{-3mm}
\begin{alignat}{2}
    c^l_t &= c^l_t \cdot (m^l_t + mg^l_t)\;.
\end{alignat}

This means, when we have a higher estimated motion, the distance metric will be increased and a smaller basis cache-rate will be selected from the codebook. As previously discussed, we also enforce a common motion-regularization in all DiT layers by computing a common motion score (\ie, $m^l_t=m_t$, $mg^l_t=mg_t$), ensuring the stability of the denoising process. We can also choose to compute motion at different frame-rates, which we ablate in our experiments. Refer to the supplementary for concrete examples of motion-score and motion-gradient (\fref{fig:supp_mograd}).

\section{Experiments}

\subsection{Implementation details}

We select multiple prominent open-source video DiTs as backbone video generation pipelines in our experiments, namely, Open-Sora-v1.2 \citep{opensora}, Open-Sora-Plan-v1.1 \citep{opensora_plan} and Latte \citep{ma2024latte}. Since we focus on inference-based latency optimizations (\ie, without any re-training), we compare AdaCache against similar methods such as $\Delta$-DiT \citep{chen2024deltadit}, T-GATE \citep{tgate} and PAB \citep{zhao2024pab}. In our main experiments, we generate 900+ videos based on standard VBench \citep{huang2023vbench} benchmark prompts at the corresponding generation settings of each baseline (\eg 480p - 2s with 30-steps in Open-Sora, 512$\times$512 - 2.7s with 150-steps in Open-Sora-Plan and 512$\times$512 - 2s with 50-steps in Latte) measuring multiple quality-complexity metrics. We report VBench average and reference-based PSNR, SSIM and LPIPS as quality metrics, and also report FLOPs, Latency (s) and Speedup as complexity metrics. Here, Latency is measured on a single A100 GPU. Unless otherwise stated, in our ablations and qualitative results, we experiment on the prompts from Open-Sora benchmark gallery, generating 720p - 2s videos with 100-steps.

\subsection{Main results}

\begin{table*}[t!]
    \scriptsize \centering
    \caption{\textbf{Quantitative evaluation of quality and latency:} Here, we compare AdaCache with other \textit{training-free} DiT acceleration methods (\eg $\Delta$-DiT \citep{chen2024deltadit}, T-GATE \citep{tgate}, PAB \citep{zhao2024pab}) on mutliple video baselines (\eg Open-Sora \citep{opensora} 480p - 2s at 30-steps, Open-Sora-Plan \citep{opensora_plan} 512$\times$512 - 2.7s at 150-steps, Latte \citep{ma2024latte} 512$\times$512 - 2s at 50-steps). 
    We measure the generation quality with VBench \citep{huang2023vbench}, PSNR, LPIPS and SSIM, while reporting complexity with FLOPs, latency and speedup (measured on a single A100 GPU). AdaCache-fast consistently shows the best speedups at a comparable or slightly-lower generation quality. AdaCache-slow gives absolute-best quality while still being faster than prior methods. Our motion-regularization significantly improves the generation quality consistently, with a minimal added-latency.
    }
    \begin{tabular}{lccccccc}
        \toprule
        Method & VBench (\%) $\uparrow$ & PSNR $\uparrow$  & LPIPS $\downarrow$ & SSIM $\uparrow$ & FLOPs (T) & Latency (s) & Speedup \\
        \midrule
        Open-Sora \citep{opensora} & 79.22 & -- & -- & -- & 3230.24 & 54.02 & 1.00$\times$  \\
        $\;+$ $\Delta$-DiT \citep{chen2024deltadit} & 78.21 & 11.91 & 0.5692 & 0.4811 & 3166.47 & -- & --  \\
        $\;+$ T-GATE \citep{tgate} & 77.61 & 15.50 & 0.3495 & 0.6760 & 2818.40 & 49.11 & 1.10$\times$ \\
        $\;+$ PAB-fast \citep{zhao2024pab} & 76.95 & 23.58 & 0.1743 & 0.8220 & 2558.25 & 40.23 & 1.34$\times$ \\
        $\;+$ {PAB}-slow \citep{zhao2024pab} & {78.51} & {27.04} & {0.0925} & {0.8847} & {2657.70} & 44.93 & 1.20$\times$ \\
        \rowcolor{row} $\;+$ AdaCache-fast & 79.39 & 24.92 & 0.0981 & 0.8375 & \textbf{1331.97} & \textbf{24.16} & \textbf{2.24$\times$} \\
        \rowcolor{row} $\;+$ AdaCache-fast (w/ MoReg) & 79.48 & 25.78 & 0.0867 & 0.8530 & 1383.66 & 25.71 & 2.10$\times$ \\
        \rowcolor{row} $\;+$ AdaCache-slow & \textbf{79.66} & \textbf{29.97} & \textbf{0.0456} & \textbf{0.9085} & 2195.50 & 37.01 & 1.46$\times$ \\
        \midrule
        
        Open-Sora-Plan \citep{opensora_plan} & 80.39 & -- & -- & -- & 12032.40 & 129.67 & 1.00$\times$  \\
        $\;+$ $\Delta$-DiT \citep{chen2024deltadit} & 77.55 & 13.85 & 0.5388 & 0.3736 & 12027.72 & -- & -- \\
        $\;+$ T-GATE \citep{tgate} & 80.15& 18.32 & 0.3066 & 0.6219 & 10663.32 & 113.75 & 1.14$\times$ \\
        $\;+$ {PAB-fast} \citep{zhao2024pab} & 71.81 & 15.47 & 0.5499 & 0.4717 & 8551.26 & 89.56 & 1.45$\times$ \\
        $\;+$ {PAB}-slow \citep{zhao2024pab} & {80.30} & {18.80} & {0.3059} & {0.6550} & {9276.57} & 98.50 & 1.32$\times$ \\
        \rowcolor{row} $\;+$ AdaCache-fast & 75.83 & 13.53 & 0.5465 & 0.4309 & \textbf{3283.60} & \textbf{35.04} & \textbf{3.70$\times$} \\
        \rowcolor{row} $\;+$ AdaCache-fast (w/ MoReg) & 79.30 & 17.69 & 0.3745 & 0.6147 & 3473.68 & 36.77 & 3.53$\times$ \\
        \rowcolor{row} $\;+$ AdaCache-slow & \textbf{80.50} & \textbf{22.98} & \textbf{0.1737} & \textbf{0.7910} & 4983.30 & 58.88 & 2.20$\times$ \\
        \midrule 
        
        Latte \citep{ma2024latte} & 77.40 & -- & -- & -- & 3439.47 & 32.45 & 1.00$\times$  \\
        $\;+$ $\Delta$-DiT \citep{chen2024deltadit} & 52.00 & 8.65 & 0.8513 & 0.1078 & 3437.33 & -- & -- \\
        $\;+$ T-GATE \citep{tgate} & 75.42 & 19.55 & 0.2612 & 0.6927 & 3059.02 & 29.23 & 1.11$\times$ \\
        $\;+$ {PAB-fast} \citep{zhao2024pab} & 73.13 & 17.16 & 0.3903 & 0.6421 & 2576.77 & 24.33 & 1.33$\times$ \\
        $\;+$ {PAB}-slow \citep{zhao2024pab} & {76.32} & {19.71} & {0.2699} & {0.7014} &  {2767.22}& 26.20 & 1.24$\times$ \\
        \rowcolor{row} $\;+$ AdaCache-fast & 76.26 & 17.70 & 0.3522 & 0.6659 & \textbf{1010.33} & \textbf{11.85} & \textbf{2.74$\times$} \\
        \rowcolor{row} $\;+$ AdaCache-fast (w/ MoReg) & 76.47 & 18.16 & 0.3222 & 0.6832 & 1187.31 & 13.20 & 2.46$\times$ \\
        \rowcolor{row} $\;+$ AdaCache-slow & \textbf{77.07} & \textbf{22.78} & \textbf{0.1737} & \textbf{0.8030} & 2023.65 & 20.35 & 1.59$\times$ \\
        \bottomrule
    \end{tabular}
    \label{tab:main}
\end{table*}

In \tref{tab:main}, we present a quantitative evaluation of quality and latency on VBench \citep{huang2023vbench} benchmark. We consider three variants of AdaCache: a slow variant, a fast variant with more speedup and the same with motion regularization. We compare with other training-free acceleration methods, showing consistently better speedups with a comparable generation quality. With Open-Sora \citep{opensora} baseline, AdaCache-slow outperforms others on all quality metrics, while giving a 1.46$\times$ speedup compared to PAB \citep{zhao2024pab} with 1.20$\times$ speedup. AdaCache-fast gives the highest acceleration of 2.24$\times$ with a slight drop in quality. AdaCache-fast (w/ MoReg) shows a clear improvement in quality compared to AdaCache-fast, validating the effectiveness of our regularization and giving a comparable speedup of 2.10$\times$. All AdaCache variants outperform even the baseline (w/o any acceleration) on VBench average quality, which aligns better with human preference compared to other reference-based metrics. Similar observations hold with the other baselines as well. With Open-Sora-Plan \citep{opensora_plan}, AdaCache shows the best speedup of 3.70$\times$ compared to the previous-best 1.45$\times$ of PAB, and the best quality with a 2.20$\times$ speedup. With Latte \citep{ma2024latte}, we gain the best speedup of 2.74$\times$ compared to prior 1.33$\times$, and the best overall quality with a 1.59$\times$ speedup.

\begin{wrapfigure}{r}{0.45\textwidth}
\centering
\vspace{-4mm}
\includegraphics[width=0.45\columnwidth]{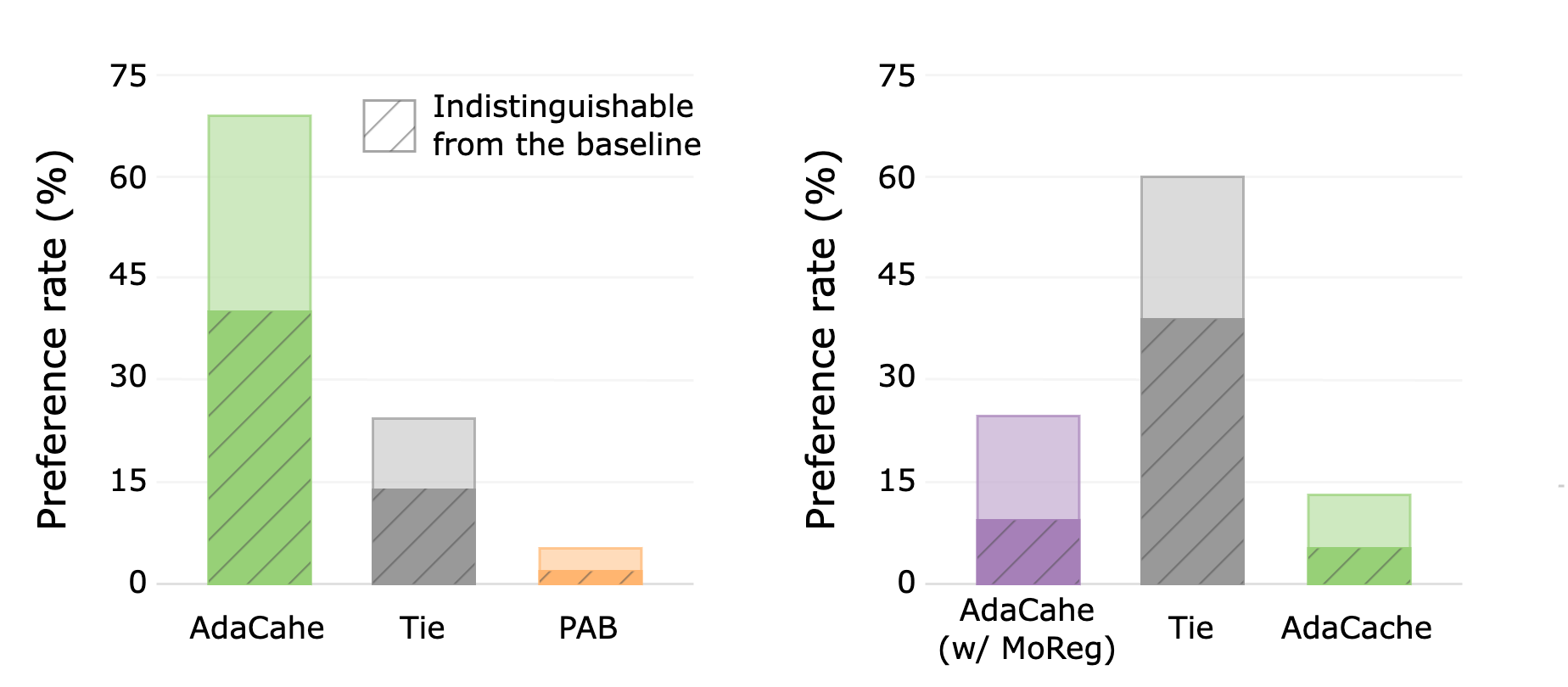}
\vspace{-5mm}
\caption{\textbf{User study:} We collect human preferences, comparing AdaCache with PAB \citep{zhao2024pab} (left) and evaluating our motion regularization (right). AdaCache 
shows a significantly-higher preference-rate over PAB at a comparable latency. Our motion- regularized variant is better-preferred, yet often tied with AdaCache in terms of perceived quality.
}
\vspace{-2mm}
\label{fig:user_study}
\end{wrapfigure}

\paragraph{User study:} Quantitative metrics on video generation quality can sometimes fall-short in aligning with the perceived visual quality. To better understand the human preference on AdaCache and its comparisons, we conduct a user study in the form of randomized A/B preference tests. Here, we create a questionnaire with 50 multiple-choice questions, each consisting of 3 variants of a single video sequence: the baseline, and two efficient generations in a randomized order (comparing either AdaCache vs. PAB \citep{zhao2024pab} at a similar speedup, or AdaCache vs. AdaCache w/ MoReg). We ask the users which efficient variant shows a better quality, and whether it is aligned with (\ie, indistinguishable from) the baseline. We collect a total of 1800 responses from 36 different users, and the results of the study are given in \fref{fig:user_study}. Between AdaCache and PAB, we see a clear win for our method (70\%) while being extremely-similar to the baseline more than half the time (41\%). Among AdaCache variants, users find these to be often tied (60\%) in-terms of perceived quality, yet still showing a better preference for motion-regularized variant (25\% vs. 14\%). This study validates the effectiveness of Adaptive Caching.

\subsection{Ablation study}

\begin{figure}[t!]
\centering
\includegraphics[width=1\columnwidth]{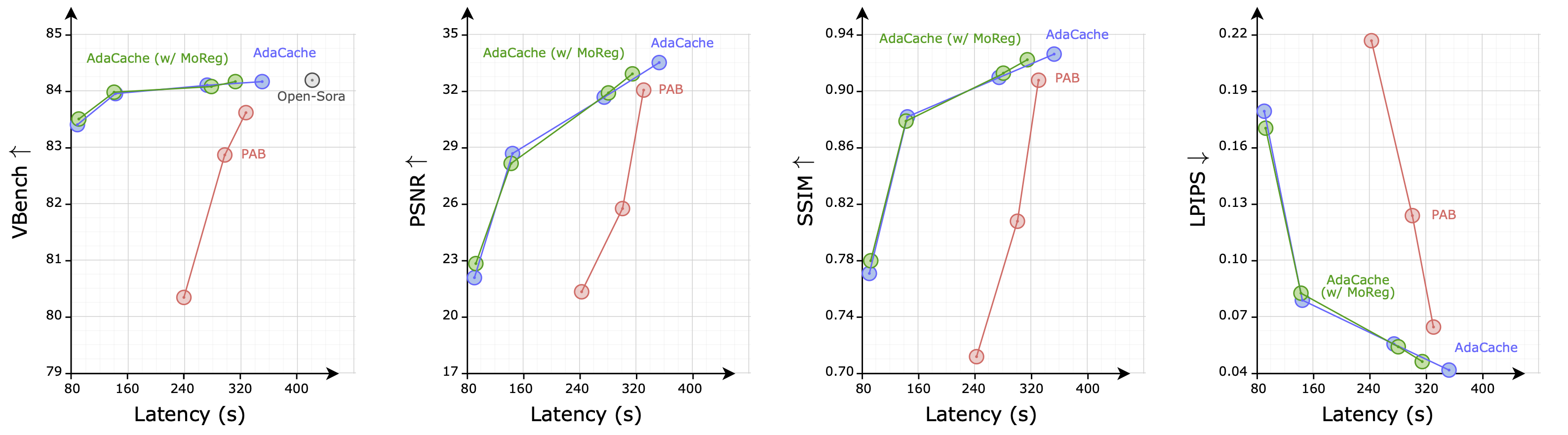}
\caption{\textbf{Quality-Latency trade-off:} We show quality vs.~latency curves for different configurations of AdaCache and PAB \citep{zhao2024pab}, with Open-Sora \citep{opensora} 720p - 2s generations. AdaCache outperforms PAB consistently, showing a more-stable performance while reducing latency. This stability is more-prominent in reference-free metric VBench \citep{huang2023vbench} compared to reference-based metrics, validating that AdaCache generations are aligned with human preference even at its fastest speeds, despite not being exactly-aligned with the reference.}
\label{fig:tradeoff}
\end{figure}

\paragraph{Quality-Latency trade-off:} In \fref{fig:tradeoff}, we compare the quality-latency trade-off of AdaCache with PAB \citep{zhao2024pab}. First, we note that AdaCache enables significantly higher reduction rates (\ie, much-smaller absolute latency) compared to PAB. Moreover, across this whole range of latency configurations, AdaCache gives a more-stable performance over PAB, on all quality metrics. Such behavior is especially evident in reference-free metric VBench \citep{huang2023vbench}, that aligns better with human preference. Even if we see a drop in reference-based scores (\eg PSNR, SSIM) at extreme reduction rates, the qualitative results suggest that the generations are still good (see \fref{fig:teaser}), despite not being aligned exactly with the reference.

\begin{figure}[t]
\centering
\includegraphics[width=1\columnwidth]{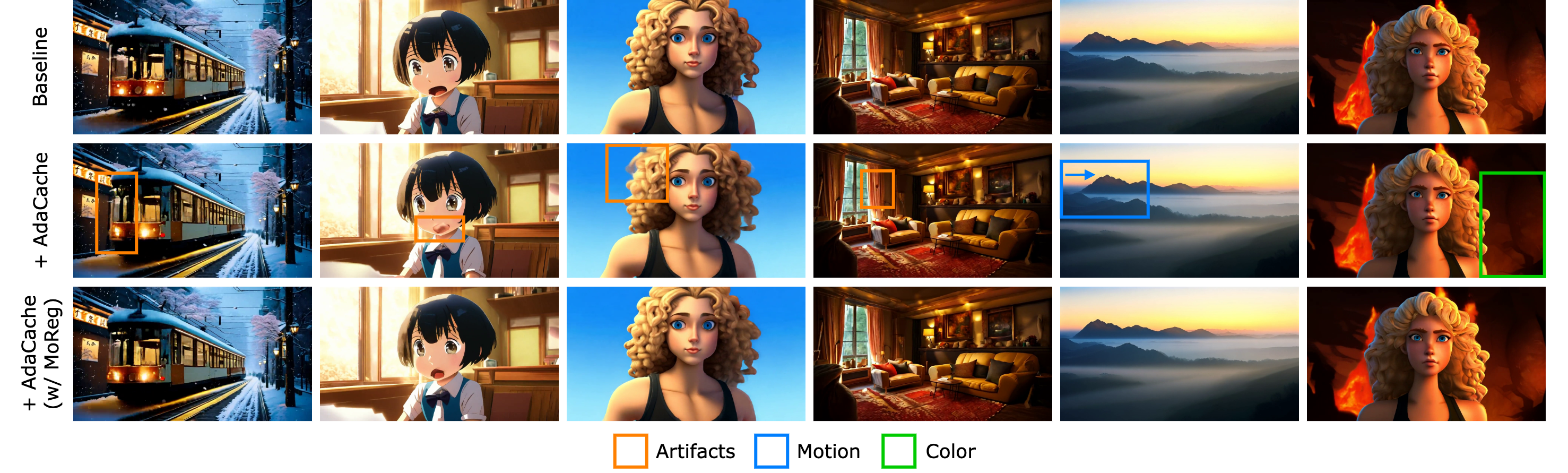}
\vspace{-5mm}
\caption{\textbf{Impact of Motion Regularization on Adaptive Caching:} We show a qualitative comparison of AdaCache and AdaCache (w/ MoReg), applied on top of Open-Sora \citep{opensora} baseline. Here, we consider generation of 720p - 2s clips at 100-steps. Despite giving a 4.7$\times$ speedup, AdaCache can also introduce some inconsistencies over time (\eg artifacts, motion, color). Motion Regularization helps avoid most of them by allocating more computations proportional to the amount of motion (while still giving a 4.5$\times$ speedup). Best-viewed with zoom-in. Prompts and more visualizations (see \fref{fig:supp_moreg}) are given in supplementary.}
\vspace{-2mm}
\label{fig:moreg}
\end{figure}

\paragraph{AdaCache with Motion Regularization:} We compare AdaCache with different versions of motion regularization in \tref{tab:ablation:adacache_variants}. Both vanilla and motion-regularized versions provide significant speedups, 4.7$\times$ and 4.5$\times$ respectively, at a comparable quality with baseline Open-Sora \citep{opensora}. Considering motion-gradient as an early-predictor of motion at latter diffusion steps helps (83.50 vs.~83.36 on VBench). We also estimate motion at different frame-rates by considering a varying step-size in frame differences, which seems to increase the latency without necessarily improving quality. Overall, we consider AdaCache (w/ MoReg) as the configuration with the best quality-latency trade-off. This improvement in quality is more-prominent in qualitative examples shown in \fref{fig:moreg}, \fref{fig:supp_moreg} and the benchmark comparison in \tref{tab:main}.

\setcounter{table}{2}
\begin{figure}[t!]
    \centering
    \includegraphics[width=0.55\columnwidth]{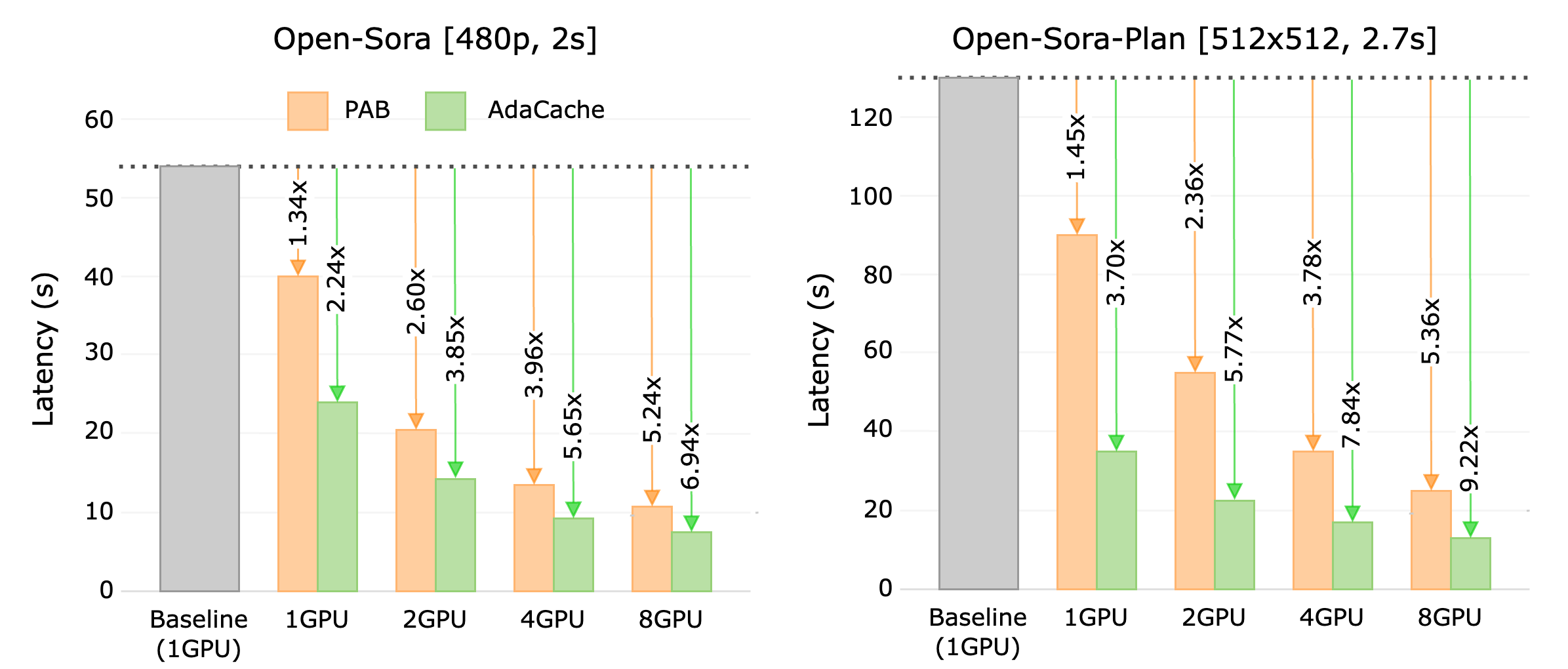}
    \quad
    \resizebox{0.42\columnwidth}{!}{
\begin{tabu}{lcccc}
\toprule
  \multirow{2}{*}{Method} &  \multicolumn{4}{c}{Latency (s) \textcolor{darkgray}{\scriptsize(speedup)}} \\
  \cmidrule{2-5}
  & 1GPU & 2GPU & 4GPU & 8GPU \\
\midrule
Open-Sora & 54.02 \textcolor{darkgray}{\scriptsize(1.00$\times$)} & 29.28 \textcolor{darkgray}{\scriptsize(1.84$\times$)} & 18.08 \textcolor{darkgray}{\scriptsize(2.99$\times$)} & 12.95 \textcolor{darkgray}{\scriptsize(4.17$\times$)} \\
 $\;+$ PAB & 40.23 \textcolor{darkgray}{\scriptsize(1.34$\times$)} & 20.77 \textcolor{darkgray}{\scriptsize(2.60$\times$)} & 13.63 \textcolor{darkgray}{\scriptsize(3.96$\times$)} & 10.31 \textcolor{darkgray}{\scriptsize(5.24$\times$)} \\
 \rowcolor{row} $\;+$ AdaCache & 24.16 \textcolor{darkgray}{\scriptsize(2.24$\times$)} & 14.02 \textcolor{darkgray}{\scriptsize(3.85$\times$)} & $\;\:$9.56 \textcolor{darkgray}{\scriptsize(5.65$\times$)} & $\;\:$7.78 \textcolor{darkgray}{\scriptsize(6.94$\times$)} \\
\midrule
Open-Sora-Plan & 129.67 \textcolor{darkgray}{\scriptsize(1.00$\times$)}$\;\:$ & 79.67 \textcolor{darkgray}{\scriptsize(1.63$\times$)} & 47.86 \textcolor{darkgray}{\scriptsize(2.71$\times$)} & 33.34 \textcolor{darkgray}{\scriptsize(3.89$\times$)} \\
 $\;+$ PAB & 89.56 \textcolor{darkgray}{\scriptsize(1.45$\times$)} & 54.87 \textcolor{darkgray}{\scriptsize(2.36$\times$)} & 34.29 \textcolor{darkgray}{\scriptsize(3.78$\times$)} & 24.21 \textcolor{darkgray}{\scriptsize(5.36$\times$)} \\
 \rowcolor{row} $\;+$ AdaCache & 35.04 \textcolor{darkgray}{\scriptsize(3.70$\times$)} & 22.49 \textcolor{darkgray}{\scriptsize(5.77$\times$)} & 16.54 \textcolor{darkgray}{\scriptsize(7.84$\times$)} & 14.07 \textcolor{darkgray}{\scriptsize(9.22$\times$)} \\
 \bottomrule
 \\\\\\\\\\\\\\\\\\\\\\\\\\
\end{tabu}}

    \vspace{-30mm}
    \captionsetup{labelformat=andtable}
    \caption{\textbf{Acceleration in multi-GPU setups:} We evaluate the speedups with varying GPU parallelization, as cached-steps can avoid communication overheads among GPUs. Here, we compare AdaCache with PAB \citep{zhao2024pab}, on baselines Open-Sora \citep{opensora} 480p - 2s generations at 30-steps and Open-Sora-Plan \citep{opensora_plan} 512$\times$512 - 2.7s generations at 150-steps. (Left) AdaCache consistently shows better acceleration over PAB in all settings. (Right) When compared with baselines of similar parallelization, the additional speedup from AdaCache increases with more GPUs. All latency measurements are on A100 GPUs.}
    \label{fig:gpu_cost}
  \end{figure}

\begin{table*}[t!]\centering
    \caption{\textbf{Ablation study:} We evaluate different design decisions of AdaCache on Open-Sora \citep{opensora} benchmark prompts, reporting VBench \citep{huang2023vbench} scores (\%), latency (s) and speedup. Here, we consider 32 videos generated with 100 diffusion steps, and use VBench custom dataset evaluation.
    }
    \vspace{-1mm}
    \subfloat[\textbf{AdaCache with Motion Regularization}: We show different variants of AdaCache. All versions achieve significant speedups compared to the baseline. AdaCache + MoReg shows a better quality 
    with a slightly-lower speedup.
    \label{tab:ablation:adacache_variants}]{
        \resizebox{0.52\columnwidth}{!}{
        \begin{tabu}{lccc}
        \toprule
          \multicolumn{1}{l}{Method}  & VBench & Latency & Speedup \\
        \midrule
        Open-Sora \citep{opensora} & 84.16 & 419.60$\;\:$ & 1.0$\times$ \\
        \midrule
         $\;+$ AdaCache & 83.40 & 89.53 & 4.7$\times$ \\
         $\;+$ AdaCache + MoReg & 83.50 & 93.50 & 4.5$\times$ \\
         $\;+$ AdaCache + MoReg (w/o grad) & 83.36 & 89.01 & 4.7$\times$ \\
         $\;+$ AdaCache + MoReg (multi-step)$\qquad$ & 83.42 & 95.65 & 4.4$\times$ \\
        \bottomrule
        \end{tabu}}}\hspace{1.5mm}
    \subfloat[\textbf{Speedups at different resolutions}: We compare AdaCache with baselines at different resolutions. It generalizes across resolutions, consistently providing a stable acceleration.
    \label{tab:ablation:resolution}]{
        \resizebox{0.43\columnwidth}{!}{
        \begin{tabu}{lcccc}
        \toprule
          \multicolumn{1}{l}{Resolution$\qquad$} & AdaCache & VBench & Latency  & Speedup \\
        \midrule
         \multirow{2}{*}{480p - 2s} & \xmark & 83.68 & 173.84$\;\:$ & 1.0$\times$ \\
         & \cmark & 83.18 & 38.52 & 4.5$\times$ \\
         \midrule
         \multirow{2}{*}{480p - 4s} & \xmark & 82.77 & 349.90$\;\:$ & 1.0$\times$ \\
         & \cmark & 82.16 & 80.16 & 4.4$\times$ \\
         \midrule
         \multirow{2}{*}{720p - 2s} & \xmark & 84.16 & 419.60$\;\:$ & 1.0$\times$ \\
         & \cmark & 83.40 & 89.53 & 4.7$\times$ \\
        \bottomrule
        \\
        \end{tabu}}}%

    \subfloat[\textbf{Cache metric}: Among distance metrics, L1/L2 have similar (and better) performance in-contrast to cosine distance.
    \label{tab:ablation:adacache_metric}]{
        \resizebox{0.21\columnwidth}{!}{
        \begin{tabu}{lcc}
        \toprule
          \multicolumn{1}{l}{Distance} & VBench  & Latency  \\
        \midrule
         L1 & 83.40 & 89.53 \\
         L2 & 83.50 & 92.70 \\
         Cosine & 83.19 & 86.74 \\
        \bottomrule
        \end{tabu}}}\hspace{1.5mm}
    \subfloat[\textbf{Cache location}: We compute the cache metric at mid-DiT, resulting in the best quality-latency trade-off.
    \label{tab:ablation:adacache_which_layer}]{
        \resizebox{0.205\columnwidth}{!}{
        \begin{tabu}{lcc}
        \toprule
          \multicolumn{1}{l}{Location} & VBench  & Latency  \\
        \midrule
         Start & 83.30 & 87.55 \\
         Mid & 83.40 & 89.53 \\
         End & 83.43 & 91.20 \\
         Multiple & 83.41 & 90.27 \\
        \bottomrule
        \\
        \end{tabu}}}\hspace{2mm}\vspace{-2mm}
    \subfloat[\textbf{Cache residual}: We consider different residual computations to estimate cache metric. Our default is Temp-attn.
    \label{tab:ablation:cache_residual}]{
        \resizebox{0.20\columnwidth}{!}{
        \begin{tabu}{lcc}
        \toprule
          \multicolumn{1}{l}{Residual} & VBench  & Latency  \\
        \midrule
         $p_{t}$ (TA) & 83.40 & 89.53 \\
         $p_{t}$ (SA) & 83.19 & 89.06 \\
         $q_{t}$ (CA) & 83.25 & 90.70 \\
         $r_{t}$ (MLP) & 83.62 & 99.72 \\
        \bottomrule
        \end{tabu}}}\hspace{1.5mm}
    \subfloat[\textbf{AdaCache Variants}: We achieve a range of speedups (and quality) by controlling the basis cache-rates in AdaCache. Our default setting is AdaCache-fast.
    \label{tab:ablation:base_rates}]{
        \resizebox{0.31\columnwidth}{!}{
        \begin{tabu}{lccc}
        \toprule
          \multicolumn{1}{l}{AdaCache} & Basis-rates & VBench  & Latency  \\
        \midrule
         Fast & \{12, 10, 8, 6, 4, 3\} & 83.40 & $\;\:$89.53 \\
         Medium & \{8, 6, 4, 2, 1\} & 83.94 & 143.87 \\
         Slow & \{2, 1\} & 84.12 & 274.30 \\
        \bottomrule
        \end{tabu}}}

    \label{tab:ablations}
    \vspace{-4mm}
\end{table*}

\paragraph{Acceleration in multi-GPU setups:} Aligned with prior work that relies on Dynamic Sequence Parallelism (DSP) \citep{zhao2024dsp} to support high-resolution long-video generation across multiple GPUs, we evaluate how AdaCache performs in such scenarios. This evaluation is relevant in the context of efficiency, as DSP incurs additional latency overheads corresponding to the communication between GPUs, and caching mechanisms can avoid such costs by re-using previous computations. We present the results of this study in \fref{fig:gpu_cost} and Table \hyperref[fig:gpu_cost]{2}. Here, we consider Open-Sora \citep{opensora} (480p - 2s at 30-steps) and Open-Sora-Plan \citep{opensora_plan} (512$\times$512 - 2.7s at 150-steps) as baselines, and compare against prior-art PAB \citep{zhao2024pab} in terms of latency meansurements on A100 GPUs. In \fref{fig:gpu_cost}, we observe that AdaCache consistently outperforms PAB with better inference speeds across all settings. In Table \hyperref[fig:gpu_cost]{2}, we further compare our method with the corresponding baselines with similar GPU parallelization. We observe that the additional speedup due to AdaCache increases with more GPUs, verifying the impact of caching on GPU communication overhead.

\paragraph{Speedups at different resolutions:} In \tref{tab:ablation:resolution}, we compare the trade-offs of AdaCache at various resolutions of video generations, namely, 480p - 2s, 480p - 4s and 720p - 2s, all at 100-steps. AdaCache provides consistent speedups across different resolutions without affecting the quality.

\paragraph{Cache metric, location and residual:} When adaptively deciding the caching schedule, we consider different metrics to compute the rate-of-change between representations, namely, L1/L2 distance or cosine distance. Among these, L1/L2 give an absolute measure which aligns better with the actual change. In contrast, cosine computes a normalized-distance, which is not a good estimate of change (\eg if the representations differ only by a scale, the distance will be zero, even though we want to have a non-zero value). This observation is verified by the results in \tref{tab:ablation:adacache_metric}. Moreover, we consider computing the cache metric at various locations (\ie, layers) in the DiT. Doing so at a single layer (\eg start, mid, end) is not significantly different from computing an aggregate over multiple-layers (see \tref{tab:ablation:adacache_which_layer}). By default, we compute the cache metric in the mid-layer as a reasonable choice without extra overheads. As for the choice of residual to be used for the cache metric computation, we resort to temporal-attention as it achieves the best trade-off (see \tref{tab:ablation:cache_residual}). 

\paragraph{AdaCache variants:} To achieve a range of speedups (and quality), we consider different basis cache-rates in our AdaCache implementation. For instance, we can have higher-speedup with a slightly-lower quality (\eg AdaCache-fast), a lower-speedup with a higher-quality (\eg AdaCache-slow), or balance both (\eg AdaCache-medium). We can conveniently control this by having corresponding basis cache-rates as shown in \tref{tab:ablation:base_rates}. By defualt, we resort to AdaCache-fast which gives the best speedups.

\subsection{Qualitative results}

\begin{figure}[t]
\centering
\includegraphics[width=1\columnwidth]{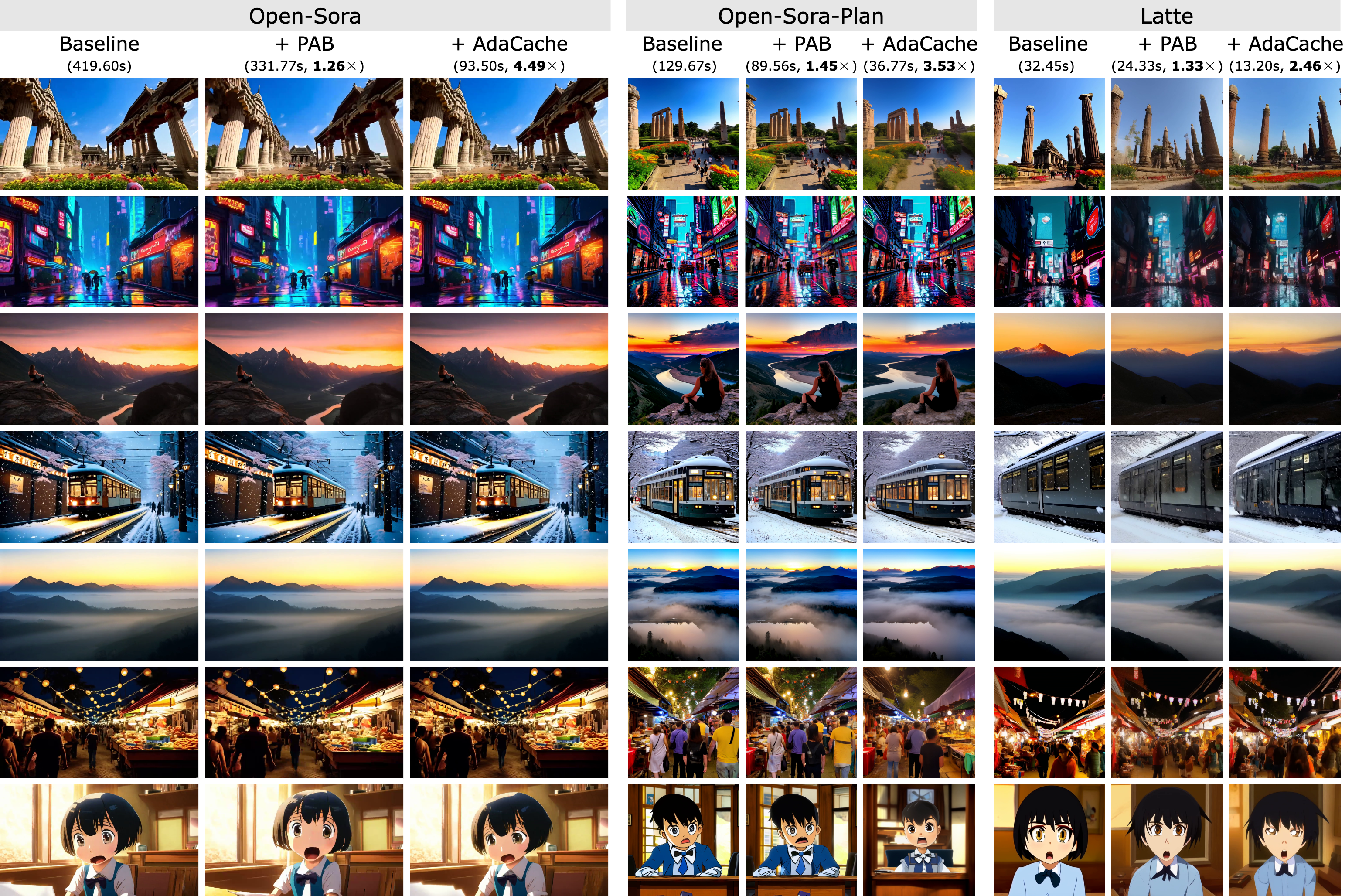}
\vspace{-6mm}
\caption{\textbf{Qualitative comparison:} We show qualitative results on multiple video-DiT baselines including Open-Sora \citep{opensora} (720p - 2s at 100-steps), Open-Sora-Plan \citep{opensora_plan} (512$\times$512 - 2.7s at 150-steps) and Latte \citep{ma2024latte} (512$\times$512 - 2s at 50-steps), while comparing against prior \textit{training-free} inference acceleration method PAB \citep{zhao2024pab}. AdaCache shows a comparable generation quality at much-faster speeds. Best-viewed with zoom-in. Prompts and additional qualitative results (\fref{fig:supp_sota}) are given in supplementary. %
}
\vspace{-3mm}
\label{fig:quali}
\end{figure}

In \fref{fig:quali}, we present qualitative results on mutliple video DiT baselines, including Open-Sora \citep{opensora}, Open-Sora-Plan \citep{opensora_plan} and Latte \citep{ma2024latte}. We compare AdaCache against each baseline and prior training-free inference acceleration method for video DiTs, PAB \citep{zhao2024pab}. Here, we consider three different configurations: 720p - 2s generations at 100-steps for Open-Sora, 512$\times$512 - 2.7s generations at 150-steps for Open-Sora-Plan, and 512$\times$512 - 2s generations at 50-steps for Latte, while considering prompts from Open-Sora gallery (see supplementary for prompt details). AdaCache shows a comparable generation quality, while having much-faster inference pipelines. In fact, it achieves 4.49$\times$ (vs. 1.26$\times$ in PAB), 3.53$\times$ (vs. 1.45$\times$ in PAB), 2.46$\times$ (vs. 1.33$\times$ in PAB) speedups respectively on the three considered baseline video DiTs. In most cases our generations are aligned well with the baseline in the pixel space. Yet this is not a strict requirement, as the denoising process can deviate considerably from that of the baseline, particularly at high caching rates. Still, AdaCache is faithful to the text prompt and is not affected by significant artifacts. Refer \fref{fig:supp_sota} for additional qualitative comparisons.

\section{Conclusion}

In this paper, we introduced Adaptive Caching (\textit{AdaCache}), a plug-and-play component that improves the the inference speed of video diffusion transformers without needing any re-training. It caches residual computations, while also devising the caching schedule dependent on each video generation. We further proposed a Motion Regularization (\textit{MoReg}) 
to utilize video information and allocate computations based on motion content, improving the quality-latency trade-off. We apply our contributions on multiple open-source video DiTs, showing comparable generation quality at a fraction of latency. We believe AdaCache is widely-applicable with minimal effort, helping democratize high-fidelity long-video generation.

\subsection*{Acknowledgements}
The authors would like to thank Ankit Khedia, Arvind Somasundaram, Dustin Johnson, Eugene Vecharynski, Ly Cao, SK Bong, and Yuzi He for the interesting discussions, and Shikun Liu for the support in setting-up the project page. The authors also appreciate the time and effort volunteered by the participants of the user study.

\subsection*{Ethics Statement}
This paper introduces a generic training-free inference acceleration mechanism for video diffusion transformers. The merits of the proposed method are evaluated on publicly-available open-source video-DiTs without being tied to a specific model or any commercial application. Consequently, the potential negative impacts of our method align with those of other video generation models and it pose no unique risk that requires special consideration. 

\subsection*{Reproducibility Statement}
This paper considers open-source video DiTs (w/ publicly-available code and pretrained-weights) in all presented experiments. As it relies on zero-shot (\ie, training-free) inference acceleration, it requires no updates to pretrained weights. All quantitative evaluations and generated videos correspond to benchmark prompts that are publicly-available. The paper details all required steps to reproduce the proposed contributions and the code is also released to the public, supporting further research on efficient video generation. 

\newpage
\renewcommand{\thetable}{A.\arabic{table}}
\renewcommand{\thefigure}{A.\arabic{figure}}
\renewcommand{\thesection}{A}
\renewcommand{\thesubsection}{A.\arabic{subsection}}
\setcounter{table}{0}
\setcounter{figure}{0}

\section{Appendix}

\begin{figure}[h]
\centering
\includegraphics[width=0.8\columnwidth]{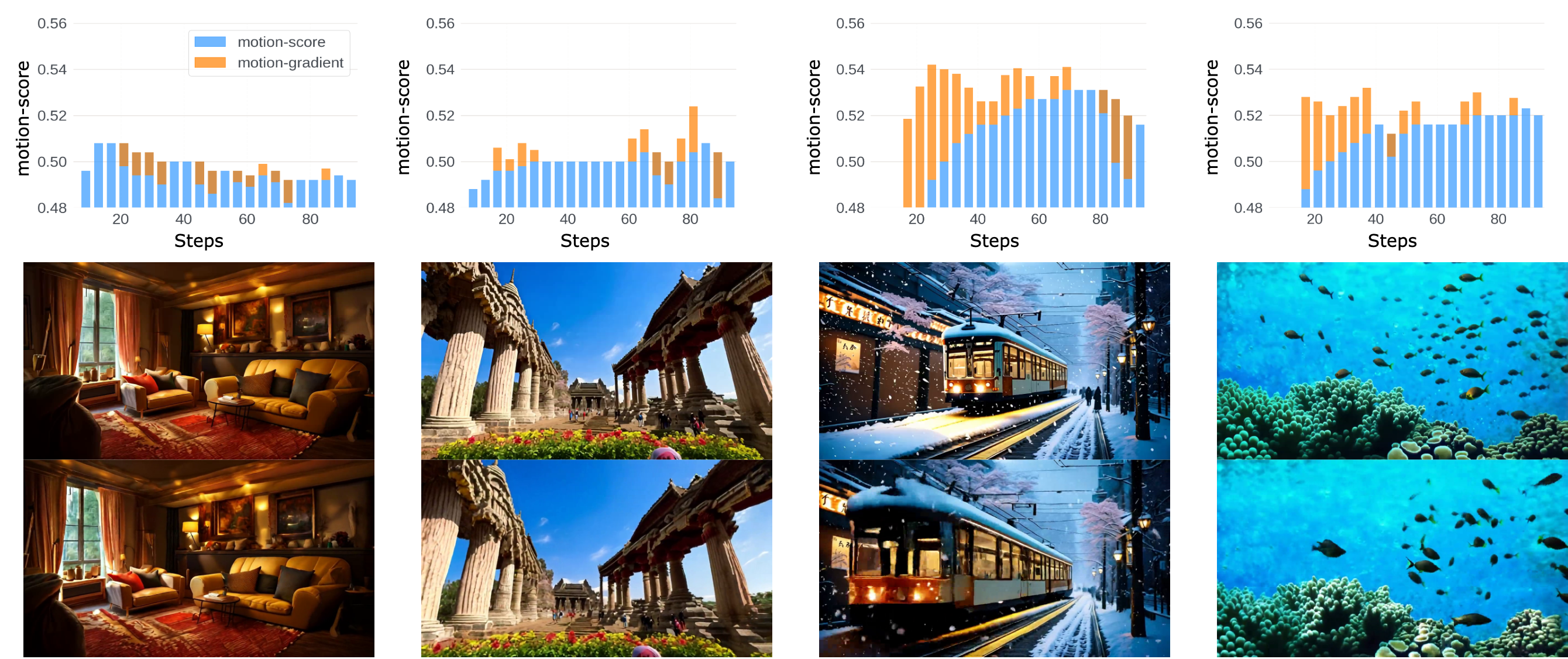}
\vspace{-1mm}
\caption{\textbf{Change in motion-score and motion-gradient across steps:} We show the histograms of Motion Regularization metrics (namely, motion-score and motion-gradient) across diffusion steps. Here, motion-score is estimated as latent frame-differences, which correlates well with the perceived motion of a given video sequence. However, it can be unreliable in early denoising steps as such latent representations are noisy. To predict the actual motion (\ie, motion in latter steps $\approx$ motion in pixel space) early, we rely on motion-gradient \textit{across diffusion steps}. Together, motion-score and motion-gradient provide a reasonable regularization. Best-viewed with zoom-in. Prompts given in supplementary.}
\vspace{-2mm}
\label{fig:supp_mograd}
\end{figure}

\subsection{Design decisions}

\paragraph{Motion-score and motion-gradient:} We rely on two metrics in our Motion regularization: namely, motion-score ($m_t$) and motion-gradient ($ mg_t$). As previously-discussed, motion-score can be unreliable particularly in early diffusion steps as it is estimated based on noisy-latents. For instance, in videos with higher motion content, our motion-score often starts small and gradually increases towards the end of diffusion process (see the two rightmost columns in \fref{fig:supp_mograd}). In slow-moving videos, motion-score can start higher can converge to a smaller value (see the leftmost column in \fref{fig:supp_mograd}). Simply put, we need a predictor of actual motion (\ie, motion in latter steps $\approx$ motion in pixel space) early in the diffusion process for a proper caching regularization. Therefore, we compute a motion-gradient across diffusion-steps, which can act as such a reasonable predictor (orange bars in \fref{fig:supp_mograd}). Together, motion-score and motion-gradient regularize the caching schedule, allocating computations based on the motion content of the video being generated.

\paragraph{Codebook of basis cache-rates:} We devise our caching schedule based on a pre-defined codebook of \textit{basis cache-rates}. It is a collection of cache-rates that is specific to a denoising schedule (\ie, \#steps), coupled with distance metric ($c_t$) thresholds for selection. Both basis cache-rates and thresholds are hyperparameters. Here, optimal thresholds may need to be tuned per video-DiT baseline, whereas the cache-rates can be adjusted depending on the required speedup (\eg AdaCache-fast, AdaCache-slow). For instance, on Open-Sora \cite{opensora} baseline, we use the codebook \texttt{\{0.08: 6, 0.16: 5, 0.24: 4, 0.32: 3, 0.40: 2, 1.00: 1\}} for AdaCache-fast in a 30-step denoising schedule, and the codebook \texttt{\{0.03: 12, 0.05: 10, 0.07: 8, 0.09: 6, 0.11: 4, 1.00: 3\}} in a 100-step schedule. For AdaCache-slow in a 30-step schedule, we use the codebook \texttt{\{0.08: 3, 0.16: 2, 0.24: 1.00: 1\}}. A specific cache-rate is selected if the distance metric is smaller than the corresponding threshold (and larger than any previous thresholds).

\subsection{Additional qualitative results}

\begin{figure}[t!]
\centering
\includegraphics[width=1\columnwidth]{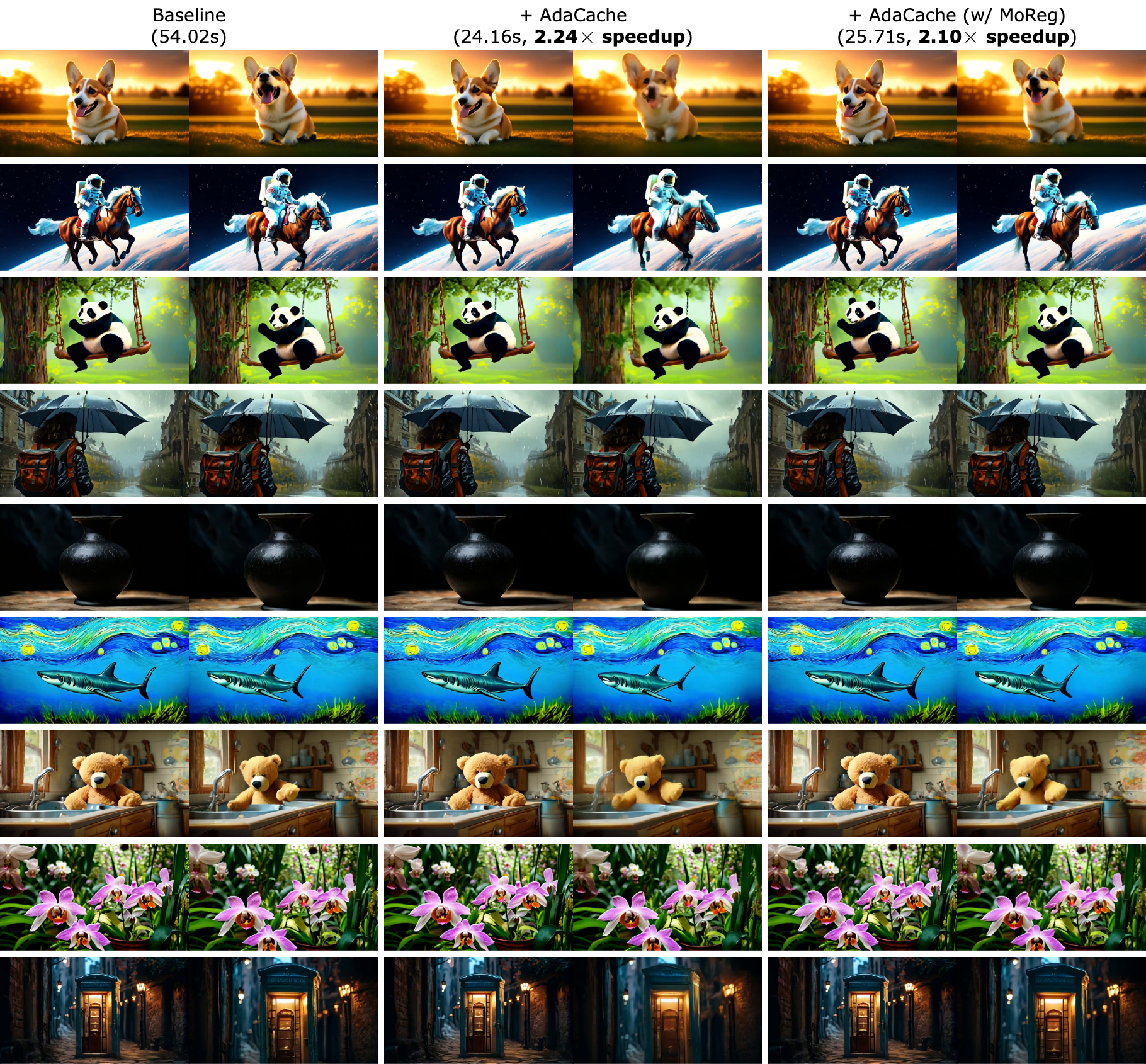}
\vspace{-6mm}
\caption{\textbf{Additional qualitative results on our Motion Regularization:} We show a qualitative comparison of AdaCache and AdaCache (w/ MoReg), applied on top of Open-Sora \citep{opensora} baseline. Here, we consider generation of 480p - 2s clips at 30-steps. Despite giving a 2.24$\times$ speedup, AdaCache can also introduce some inconsistencies over time. Our Motion Regularization helps avoid most of them by allocating computations proportional to the amount of motion (still giving a 2.10$\times$ speedup). Best-viewed with zoom-in. Prompts given in supplementary.
}
\label{fig:supp_moreg}
\end{figure}

\begin{figure}[t!]
\centering
\includegraphics[width=1\columnwidth]{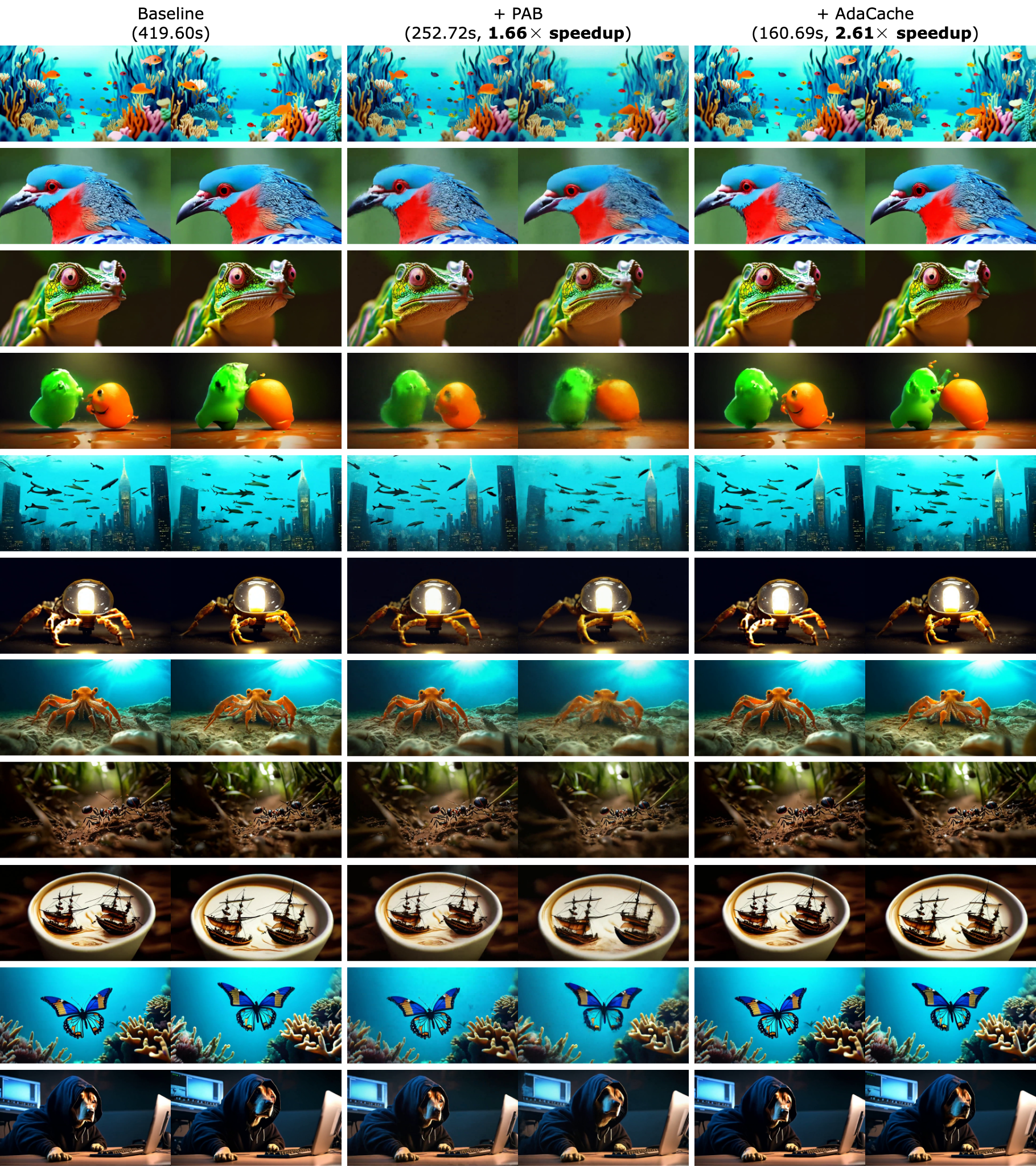}
\vspace{-6mm}
\caption{\textbf{Additional qualitative comparisons with prior-art:} We show qualitative comparisons with prior-art on baseline Open-Sora \citep{opensora} (720p - 2s at 100-steps). Here, we evaluate against prior \textit{training-free} inference acceleration method PAB \citep{zhao2024pab} at a comparable speedup. AdaCache consistently shows a better generation quality. Best-viewed with zoom-in. Prompts given in supplementary. %
}
\label{fig:supp_sota}
\end{figure}

In \fref{fig:supp_moreg}, we provide additional qualitative results, comparing AdaCache and AdaCache (w/ MoReg) with a baseline Open-Sora \citep{opensora}. Here, we consider 480p - 2s video generations at 30-steps, based on a few VBench \citep{huang2023vbench} prompts. Both versions with and without motion regularization achieve comparable speedups (2.10$\times$ and 2.24$\times$, respectively). Yet, MoReg helps stabilize the generation quality--- especially towards the end-of-sequence in long-videos--- by allocating computations proportional to the amount of motion. The generations with motion regularization also follow the corresponding baseline generations more-faithfully. In \fref{fig:supp_sota}, we present additional qualitative comparisons with prior-art at a comparable inference speedup. Here, we consider 720p - 2s video generations at 100-steps, based on a few Sora \citep{brooks2024sora} prompts. Our comparison includes PAB \citep{zhao2024pab}: another training-free video-DiT acceleration method. AdaCache consistently shows a better generation quality at a 2.61$\times$ speedup, compared to PAB, even at a 1.66$\times$ speedup. This behavior is also observed in \fref{fig:tradeoff}, as the generation quality of PAB degrades quickly at faster speeds.

\subsection{Text prompts used in qualitative examples}

In this subsection, we provide all the prompts used to generate the qualitative results shown in the paper. They consist of prompts from multiple sources including Open-Sora \citep{opensora} gallery, VBench \citep{huang2023vbench} benchmark and Sora \citep{brooks2024sora}, all of which are publicly-available.

Text prompts corresponding to the video generations in \fref{fig:teaser}:
\begin{itemize}
    \item \texttt{\small A Japanese tram glides through the snowy streets of a city, its sleek design cutting through the falling snowflakes with grace. The tram's illuminated windows cast a warm glow onto the snowy surroundings, creating a cozy atmosphere inside. Snowflakes dance in the air, swirling around the tram as it moves along its tracks. Outside, the city is blanketed in a layer of snow, transforming familiar streets into a winter wonderland. Cherry blossom trees, now bare, stand quietly along the tram tracks, their branches dusted with snow. People hurry along the sidewalks, bundled up against the cold, while the tram's bell rings softly, announcing its arrival at each stop.}
    \item \texttt{\small a picturesque scene of a tranquil beach at dawn. the sky is painted in soft pastel hues of pink and orange, reflecting on the calm, crystal-clear water. gentle waves lap against the sandy shore, where a lone seashell lies near the water's edge. the horizon is dotted with distant, low-lying clouds, adding depth to the serene atmosphere. the overall mood of the video is peaceful and meditative, with no text or additional objects present. the focus is on the natural beauty and calmness of the beach, captured in a steady, wide shot.}
    \item \texttt{\small a bustling night market scene with vibrant stalls on either side selling food and various goods. the camera follows a person walking through the crowded, narrow alley. string lights hang overhead, casting a warm, festive glow. people of all ages are talking, browsing, and eating, creating an atmosphere full of lively energy. occasional close-ups capture the details of freshly cooked dishes and colorful merchandise. the video is dynamic with a mixture of wide shots and close-ups, capturing the essence of the night market without any text or sound.}
    \item \texttt{\small a dynamic aerial shot showcasing various landscapes. the sequence begins with a sweeping view over a dense, green forest, transitioning smoothly to reveal a winding river cutting through a valley. next, the camera rises to capture a panoramic view of a mountain range, the peaks dusted with snow. the shot shifts to a coastal scene, where waves crash against rugged cliffs under a partly cloudy sky. finally, the aerial view ends over a bustling cityscape, with skyscrapers and streets filled with motion and life. the video does not contain any text or additional overlays.}
    \item \texttt{\small a cozy living room scene with a christmas tree in the corner adorned with colorful ornaments and twinkling lights. a fireplace with a gentle flame is situated across from a plush red sofa, which has a few wrapped presents placed beside it. a window to the left reveals a snowy landscape outside, enhancing the festive atmosphere. the camera slowly pans from the window to the fireplace, capturing the warmth and tranquility of the room. the soft glow from the tree lights and the fire illuminates the room, casting a comforting ambiance. there are no people or text in the video, focusing purely on the holiday decor and cozy setting.}
\end{itemize}

Text prompts corresponding to new video generations in \fref{fig:step_change}:
\begin{itemize}
    \item \texttt{\small a breathtaking aerial view of a river meandering through a lush green landscape. the river, appearing as a dark ribbon, cuts through the verdant fields and hills, reflecting the soft light of the pinkish-orange sky. the sky, painted in hues of pink and orange, suggests the time of day to be either sunrise or sunset. the landscape is dotted with trees and bushes, adding to the natural beauty of the scene. the perspective of the video is from above, providing a bird's eye view of the river and the surrounding landscape. the colors , the river, the landscape, and the sky all come together to create a serene and picturesque scene.}
    \item \texttt{\small A cozy living room, surrounded by soft cushions and warm lighting. Describe the scene in vivid detail, capturing the feeling of comfort and relaxation.}
    \item \texttt{\small a nighttime scene in a bustling city filled with neon lights and futuristic architecture. the streets are crowded with people, some dressed in high-tech attire and others in casual cyberpunk fashion. holographic advertisements and signs illuminate the area in vibrant colors, casting a glow on the buildings and streets. futuristic vehicles and motorcycles are speeding by, adding to the city's dynamic atmosphere. in the background, towering skyscrapers with intricate designs stretch into the night sky. the scene is filled with energy, capturing the essence of a cyberpunk world.}
    \item \texttt{\small a close-up shot of a vibrant coral reef underwater. various colorful fish swim leisurely around the corals, creating a lively scene. the lighting is natural and slightly subdued, emphasizing the deep-sea environment. soft waves ripple across the view, occasionally bringing small bubbles into the frame. the background fades into a darker blue, suggesting deeper waters beyond. there are no texts or human-made objects visible in the video.}
    
    \item \texttt{\small a neon-lit cityscape at night, featuring towering skyscrapers and crowded streets. the streets are bustling with people wearing futuristic attire, and vehicles hover above in organized traffic lanes. holographic advertisements are projected onto buildings, illuminating the scene with vivid colors. a light rain adds a reflective sheen to the ground, enhancing the cyberpunk atmosphere. the camera pans slowly through the scene, capturing the energy and technological advancements of the city. the video does not contain any text or additional objects.}
    \item \texttt{\small a breathtaking view of a mountainous landscape at sunset. the sky is painted with hues of orange and pink, casting a warm glow over the scene. the mountains, bathed in the soft light, rise majestically in the background, their peaks reaching towards the sky. in the foreground, a woman is seated on a rocky outcrop, her body relaxed as she takes in the vie w. she is dressed in a black dress and boots, her attire contrasting with the natural surroundings. her position on the rock provides a vantage point over a river that meanders through the valley below. the river, a ribbon of blue, winds its way through the landscape, adding a dynamic element to the scene. the woman's gaze is directed towards the river, suggesting a sense of contemplation or admiration for the beauty of nature. the video is taken from a high angle, looking down on the woman and the landscape. this perspective enhances the sense of depth and scale in the image, emphasizing the vastness of the mountains and the river.}
    \item \texttt{\small an animated scene featuring a young girl with short black hair and a bow tie, seated at a wooden desk in a warmly lit room. natural light filters through a window, illuminating the girl's wide eyes and open mouth, conveying a sense of surprise or shock. she is dressed in a blue shirt with a white collar and dark vest. the room's inviting atmosphere is complemented by wooden furniture and a framed picture on the wall. the animation style is reminiscent of japanese anime, characterized by vibrant colors and expressive character designs.}
\end{itemize}

Text prompts corresponding to new video generations in \fref{fig:moreg}:
\begin{itemize}
    \item \texttt{\small a breathtaking aerial view of a misty mountain landscape at sunrise. the sun is just beginning to peek over the horizon, casting a warm glow on the scene. the mountains, blanketed in a layer of fog, rise majestically in the background. the mist is so dense that it obscures the peaks of the mountains, adding a sense of mystery to the scene. in the foregro und, a river winds its way through the landscape, its path marked by the dense fog. the river appears calm, its surface undisturbed by the early morning chill. the colors in the video are predominantly cool, with the blue of the sky and the green of the trees contrasting with the warm orange of the sunrise. the video is taken from a high vantage point, p roviding a bird's eye view of the landscape. this perspective allows for a comprehensive view of the mountains and the river, as well as the fog that envelops them. the video doe s not contain any text or human activity, focusing solely on the natural beauty of the landscape. the relative positions of the objects suggest a vast, untouched wilderness.}
    \item \texttt{\small a 3d rendering of a female character with curly blonde hair and striking blue eyes. she is wearing a black tank top and is standing in front of a fiery backdrop. the character is looking off to the side with a serious expression on her face. the background features a fiery orange and red color scheme, suggesting a volcanic or fiery environment. the lighting in the scene is dramatic, with the character's face illuminated by a soft light that contrasts with the intense colors of the background. there are no texts or other objects in the image. the style of the image is realistic with a high level of detail, indicative of a high-quality 3d rendering.}
\end{itemize}

Text prompts corresponding to new video generations in \fref{fig:30v100}:
\begin{itemize}
    \item \texttt{\small a realistic 3d rendering of a female character with curly blonde hair and blue eyes. she is wearing a black tank top and has a neutral expression while facing the camera directly. the background is a plain blue sky, and the scene is devoid of any other objects or text. the character is detailed, with realistic textures and lighting, suitable for a video game or high-quality animation. there is no movement or additional action in the video. the focus is entirely on the character's appearance and realistic rendering.}
\end{itemize}

Text prompts corresponding to new video generations in \fref{fig:quali}:
\begin{itemize}
    \item \texttt{\small a scenic shot of a historical landmark. the landmark is an ancient temple with tall stone columns and intricate carvings. the surrounding area is lush with greenery and vibrant flowers. the sky above is clear and blue, with the sun casting a warm glow over the scene. tourists can be seen walking around, taking pictures and admiring the architecture. there is no text or additional objects in the video.}
    \item \texttt{\small a vibrant cyberpunk street scene at night. neon signs and holographic advertisements illuminate the narrow street, casting colorful reflections on the rain-slicked pavement. various characters, dressed in futuristic attire, move along the sidewalks while robotic street vendors sell their wares. towering skyscrapers with glowing windows dominate the background, creating a sense of depth. the camera takes a wide-angle perspective, capturing the bustling and lively atmosphere of the cyberpunk cityscape. there are no texts or other objects outside of the described scene.}
\end{itemize}

Text prompts corresponding to new video generations in \fref{fig:supp_moreg}:
\begin{itemize}
    \item \texttt{\small A cute happy Corgi playing in park, sunset, surrealism style}
    \item \texttt{\small An astronaut is riding a horse in the space in a photorealistic style.}
    \item \texttt{\small A panda playing on a swing set}
    \item \texttt{\small a backpack and an umbrella}
    \item \texttt{\small a black vase}
    \item \texttt{\small a shark is swimming in the ocean, Van Gogh style}
    \item \texttt{\small A teddy bear washing the dishes}
    \item \texttt{\small A tranquil tableau of a peaceful orchid garden showcased a variety of delicate blooms}
    \item \texttt{\small A tranquil tableau of the phone booth was tucked away in a quiet alley}
\end{itemize}

Text prompts corresponding to new video generations in \fref{fig:supp_sota}:
\begin{itemize}
    \item \texttt{\small A gorgeously rendered papercraft world of a coral reef, rife with colorful fish and sea creatures.}
    \item \texttt{\small This close-up shot of a Victoria crowned pigeon showcases its striking blue plumage and red chest. Its crest is made of delicate, lacy feathers, while its eye is a striking red color. The bird’s head is tilted slightly to the side, giving the impression of it looking regal and majestic. The background is blurred, drawing attention to the bird’s striking appearance.}
    \item \texttt{\small This close-up shot of a chameleon showcases its striking color changing capabilities. The background is blurred, drawing attention to the animal’s striking appearance.}
    \item \texttt{\small a green blob and an orange blob are in love and dancing together}
    \item \texttt{\small New York City submerged like Atlantis. Fish, whales, sea turtles and sharks swim through the streets of New York.}
    \item \texttt{\small nighttime footage of a hermit crab using an incandescent lightbulb as its shell}
    \item \texttt{\small A large orange octopus is seen resting on the bottom of the ocean floor, blending in with the sandy and rocky terrain. Its tentacles are spread out around its body, and its eyes are closed. The octopus is unaware of a king crab that is crawling towards it from behind a rock, its claws raised and ready to attack. The crab is brown and spiny, with long legs and antennae. The scene is captured from a wide angle, showing the vastness and depth of the ocean. The water is clear and blue, with rays of sunlight filtering through. The shot is sharp and crisp, with a high dynamic range. The octopus and the crab are in focus, while the background is slightly blurred, creating a depth of field effect.}
    \item \texttt{\small A low to the ground camera closely following ants in the jungle down into the ground into their world.}
    \item \texttt{\small Photorealistic closeup video of two pirate ships battling each other as they sail inside a cup of coffee.}
    \item \texttt{\small a photorealistic video of a butterfly that can swim navigating underwater through a beautiful coral reef}
    \item \texttt{\small A computer hacker labrador retreiver wearing a black hooded sweatshirt sitting in front of the computer with the glare of the screen emanating on the dog's face as he types very quickly.}
\end{itemize}

\newpage
\bibliographystyle{assets/plainnat}
\bibliography{egbib}

\end{document}